\documentclass{article}

\usepackage{amssymb} 
\usepackage{amsmath}
 \usepackage[eandd, final, nonatbib]{neurips_2026}
\usepackage{natbib}
\usepackage[utf8]{inputenc} % allow utf-8 input
\usepackage[T1]{fontenc}    % use 8-bit T1 fonts
\usepackage{hyperref}       % hyperlinks
\usepackage{url}            % simple URL typesetting
\usepackage{booktabs}       % professional-quality tables
\usepackage{amsfonts}       % blackboard math symbols
\usepackage{nicefrac}       % compact symbols for 1/2, etc.
\usepackage{microtype}      % microtypography
\usepackage{xcolor}         % colors
\usepackage{command/tags}
\usepackage[utf8]{inputenc} % allow utf-8 input
\usepackage[T1]{fontenc}    % use 8-bit T1 fonts
\usepackage{hyperref}       % hyperlinks
\usepackage{url}            % simple URL typesetting
\usepackage{booktabs}       % professional-quality tables
\usepackage{amsfonts}       % blackboard math symbols
\usepackage{nicefrac}       % compact symbols for 1/2, etc.
\usepackage{microtype}      % microtypography
\usepackage{xcolor}         % colors

\usepackage{amssymb}
\usepackage{colortbl}
\usepackage{wrapfig}
\usepackage{graphicx}
\usepackage{xspace}
\usepackage{multirow}
\usepackage{makecell}
\usepackage{nicematrix}
\usepackage[normalem]{ulem}
\usepackage{listings}
\usepackage{comment}
\usepackage{subfigure}
\usepackage{caption}
\usepackage{rotate}
\usepackage{float}
\usepackage{booktabs} 
\usepackage{cleveref}
\usepackage{hyperref}
\usepackage{algorithmic}
\usepackage[linesnumbered,ruled,vlined]{algorithm2e}
\usepackage[bottom]{footmisc}
\usepackage{tcolorbox}
\usepackage{tabularx}
\usepackage{arydshln}
\usepackage{multirow}

\NewDocumentCommand{\rot}{O{45} O{1em} m}{\makebox[#2][l]{\rotatebox{#1}{#3}}}%

\definecolor{babypink}{rgb}{0.96, 0.76, 0.76}
\definecolor{coralpink}{rgb}{0.97, 0.51, 0.47}
\definecolor{aqua}{rgb}{0.0, 1.0, 1.0}
\definecolor{columbiablue}{rgb}{0.61, 0.87, 1.0}
\definecolor{cyan}{RGB}{222, 255, 255}
\definecolor{turquoiseblue}{rgb}{0.0, 1.0, 0.94}
\definecolor{realcyan}{RGB}{0, 200, 200}

\hypersetup{
    colorlinks = true,
    linkbordercolor = {white},
    linkcolor = blue!60!black,
    citecolor = blue!60!black,
    urlcolor = blue!60!black
}

\newcommand{\huggingface}{\raisebox{-1.5pt}{\includegraphics[height=1.05em]{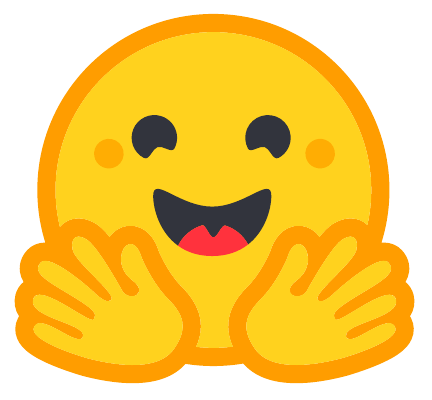}}\xspace}
\newcommand{\github}{\raisebox{-1.5pt}{\includegraphics[height=1.05em]{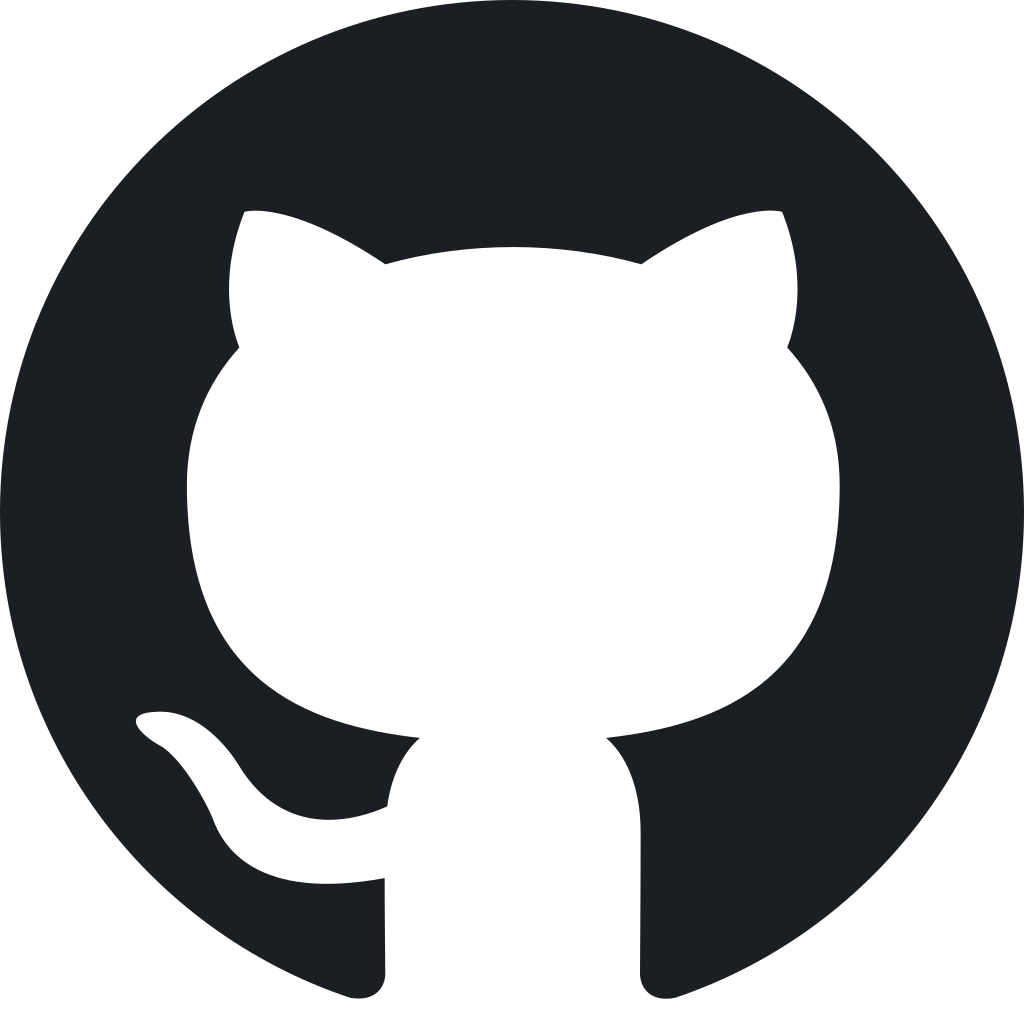}}\xspace}

\newcommand{\stride}{\textsc{Stride}}
\newcommand{\stridebench}{\textsc{Stride-Bench}}

\usepackage{xcolor}
\usepackage{soul} % for highlighting
\usepackage{listings}

\usepackage{graphicx}   % for \resizebox
\usepackage{pifont}     % for \ding

\usepackage{multirow}

\usepackage{tabularx}
\usepackage{array}

\usepackage{listings}

\lstdefinestyle{promptstyle}{
  basicstyle=\scriptsize\ttfamily,
  breaklines=true,
  breakatwhitespace=false,
  columns=fullflexible,
  keepspaces=true,
  showstringspaces=false,
  frame=single,
  xleftmargin=0pt,
  xrightmargin=0pt,
  linewidth=\linewidth
}

\title{\stride{}: Automated Evaluation of Text-to-Trajectory Alignment across Diverse Contexts}

\author{
    \textbf{Wanchun Ni$^{1}$} \quad
    \textbf{Tao Qi$^{2}$} \quad
    \textbf{Leonel Aguilar$^{1}$} \quad
    \textbf{Marlene Wagner$^{1}$} \vspace{0.1em} \\
    \textbf{Jiugeng Sun$^{1}$} \quad
    \textbf{Verena Zimmermann$^{1}$} \quad
    \textbf{Mennatallah El-Assady$^{1}$}
    \vspace{0.4em} \\
    \textnormal{$^{1}$ETH Zurich \quad $^{2}$Beijing University of Posts and Telecommunications}
    \vspace{0.4em} \\
    \texttt{wanchun.ni@inf.ethz.ch}
    \vspace{0.4em} \\
    \github \textbf{Code}: \url{https://github.com/ETH-IVIA-Lab/STRIDE-Bench}
    \vspace{0.1em} \\
    \huggingface\textbf{Dataset}:~\texttt{\href{https://huggingface.co/datasets/eth-ivia-lab/STRIDE-Bench}{eth-ivia-lab/STRIDE-Bench}}
}

\usepackage{microtype}
\begin{document}

\maketitle

\begin{abstract}
Language-conditioned trajectory generation is here, but its evaluation has not kept pace. Existing pedestrian trajectory metrics primarily compare trajectories with real-world human data. It does not scale to text-to-trajectory generation across diverse contexts, as collecting human trajectories for every scenario is costly and infeasible. Moreover, pedestrian behavior is heterogeneous and context-dependent, with no single metric as the correct answer, and current evaluation frameworks are not transferable to this domain. These challenges make scalable, reliable evaluation difficult.
We introduce \stride{}, the \textbf{first} framework for evaluating context alignment between scenario descriptions and pedestrian trajectories. \stride{} addresses these challenges through three design choices.
First, we derive our \protocol{} evaluation protocol from sociological theories to define a verifiable evaluation space.
Second, it decomposes high-level context into scenario-adaptive behavioral questions. Third, every question is resolved against a deterministic measurement tool library that yields reproducible answers. Together, \stride{} enables verifiable, automated, and scalable evaluation across diverse contexts without requiring human trajectory data.
We instantiate \stride{} in the crowd domain as \stridebench{}, comprising 1K scenarios, 6K behavioral questions, 11K measurements with calibrated expected answers across 30 real-world maps. 
Comprehensive human validations show that \stridebench{} is strongly consistent with human behavior and judgment, achieving 80\% human agreement. We further evaluate several text-to-trajectory models, finding limited context-alignment capability and persistent challenges in fine-grained context conditioning. 
We believe that our \stride{} framework provides a first step toward principled evaluation of context-aligned pedestrian trajectory generation. 
\end{abstract}

% \vspace{-1em}
\section{Introduction}
\vspace{-0.5em}

% In robotics, text instructions now drive motion synthesis that executes high-level user intent~\cite{zitkovich2023rt, li2025language}. In autonomous driving and traffic simulation, free-text scenario descriptions are used to condition vehicle trajectories for planning and safety-critical testing~\cite{song2025vl, song2024socially}. 
% Applications such as urban planning, public safety, and event design require generating pedestrian trajectories under specified conditions, but existing datasets mostly capture routine behavior and under-represent these cases.
% Large language models (LLMs) have reshaped trajectory generation across domains. 

% Pedestrian trajectory generation has followed this trend: Text-Crowd combines text and image diffusion to generate crowd scenes~\cite{ji2024text}, LMTraj-ZERO casts prediction as zero-shot LLM inference~\cite{bae2024lmtrajectory}, and a growing body of work conditions pedestrian generation on textual or symbolic behavioral specifications~\cite{cao2024crowdmogen, wei2024chatdyn, panayiotou2025gen, yu2025socialgen, kim2025guidecotgoaldrivenuserinformeddynamic}. Pedestrian prediction is also shifting toward plausible synthesis in unseen scenarios beyond the training distribution~\cite{dong2024recurrent}.

Generating realistic human trajectories is a fundamental problem for modeling how people move, interact, and respond to their surroundings. It supports applications from urban planning to embodied-agent development, where pedestrian trajectories help assess public-space accessibility and learn human-aware behavior before real-world deployment. In these settings, trajectories are not generated in isolation: they must align with the scenario across individual, group-level behavior, and the environment. For example, on the same transit-station map, pedestrians in a routine commute should follow accessible corridors at normal walking speeds, form bidirectional flows, and avoid obstacles; in an emergency evacuation, they should move faster toward exits and form directed outflows. This motivates trajectory generation that should be aligned with scenario context. Recent advances in large language models, with their capacity to interpret rich natural-language instructions, have further reshaped this landscape: Text-Crowd combines text and image diffusion to generate crowd scenes~\cite{ji2024text}, LMTraj-ZERO casts prediction as zero-shot LLM inference~\cite{bae2024lmtrajectory}, and a growing body of work conditions pedestrian generation on textual~\cite{cao2024crowdmogen, wei2024chatdyn} or symbolic behavioral~\cite{panayiotou2025gen, yu2025socialgen} specifications.

\textbf{Research Gap.}
The evaluation of text-conditioned pedestrian trajectory generation, however, remains underdeveloped. Existing evaluation protocols still rely heavily on real-world human trajectory datasets: assessing a model on a given context typically requires first collecting human trajectories for that scenario, and then comparing the consistency between the generated trajectories and the real-world ones.
However, the collecting trajectory human data for targeted contexts is highly expensive and sometimes even impractical, particularly for rare situations such as evacuations and violence. 
For example, widely used pedestrian trajectory datasets, including ETH/UCY, SDD, and TrajNet++~\cite{pellegrini2009you, lerner2007crowds, robicquet2016learning, Kothari2020HumanTF}, predominantly capture routine behaviors, leaving a long tail of scenarios under-represented. Given this scenario diversity and the scarcity of corresponding human trajectory data, human-data-based evaluation does not scale. An automatic evaluation framework that does not depend on real-world trajectories for every target scenario needs further study.

\textbf{Challenges.}
In fact, evaluating text-conditioned pedestrian trajectories is not a trivial task since it is challenging to transfer existing evaluation frameworks for content generation in such scenario. Specifically, those approaches can be broadly grouped into three categories: ground-truth-based metrics, LLM-as-judge, and human evaluation. However, in a given context, humans may exhibit many plausible behaviors. Ground-truth-based metrics may not generalize into those scenarios, thus giving a wrong judgment. 
Besides, LLM-based evaluation can be unstable and sensitive to prompts, runs, and model versions. Moreover, human evaluation, while informative, is expensive, slow, and difficult to scale across diverse scenarios. Therefore, a verifiable, reliable, automatic, and scalable evaluation framework is a prerequisite for meaningful progress in this area.

\textbf{Our approach.}
We introduce \stride{}, a framework for text-to-trajectory alignment evaluation via behavioral decomposition and structured verification. Instead of applying LLM-as-judge directly to human behaviors, \stride{} decomposes each scenario context into fine-grained behavioral questions and verifies them through structured trajectory measurements. The framework consists of three components: (i) a social-science-grounded five-axis protocol, \protocol{} (\textbf{V}elocity, \textbf{R}ealism, \textbf{D}irection, \textbf{S}patial, \textbf{T}emporal), which defines a common observable physical behavioral evaluation space in individual, group, and environment layers; (ii) scenario-adaptive behavioral questions decomposed from text descriptions under the guidance of the protocol; and (iii) structured verification of these questions. We build a \textbf{D}eterministic \textbf{M}easurement \textbf{T}ool (DMT) library in which each function computes an exact trajectory statistic. The resulting measurements determine whether generated trajectories match scenario-specific expected answers. By separating semantic curation from numerical verification, \stride{} provides a reproducible and interpretable framework for evaluating text-to-trajectory alignment. We instantiate \stride{} in the crowd setting and release \stridebench{}, a benchmark for context-aligned crowd trajectory generation.  We use models from the GPT family to construct \stridebench{}, which contains 1k scenario descriptions across 11 crowd categories, 6k behavioral questions, and 11k measurements over 30 real-world maps.

Human annotation studies show that \stridebench{} aligns well with human judgment. We validate \stridebench{} through three analyses: (i) evaluating real human trajectories, which are recognized as highly context-aligned with a \stride{} score of 0.94; (ii) measuring human agreement with benchmark answers, showing substantial consistency with human judgments; and (iii) probing SOTA LLMs, showing stable benchmark answers across different generators. We further evaluate several pedestrian trajectory generation models. Results show that (i) \stride{} discriminates context alignment across models and scales; (ii) current language-conditioned models still have substantial room for improvement, with the best baseline achieving a \stride{} score of 0.64 and exhibiting sensitivity to input configurations; and (iii) \stride{}’s fine-grained, traceable structure helps identify underperforming behavioral dimensions and inform model development.

% In summary,
% we introduce \stride{}, the first framework for evaluating text-to-trajectory alignment with behavioral decomposition and structure verification, and instantiate it as \stridebench{}, a reliable benchmark validated against human trajectories and judgments. Using \stridebench{}, we evaluate representative models and reveal substantial gaps in context alignment. As an initial step toward principled evaluation, \stride{} highlights the progress still needed to reach human-level alignment.

% \vspace{-1em}
\section{Related Work}
\vspace{-0.5em}

\paragraph{Language-conditioned trajectory generation.}
Natural language has emerged as a control interface for trajectory modeling in robotics, traffic simulation, and autonomous driving, where text specifies motion intent, scene dynamics, or high-level plans~\cite{kambara2026lilac,bamani2025speech,huang2025zlatte,tan2023language,xia2024language,wei2024chatdyn,chang2025langtraj,pan2024vlp,li2025generative,hwang2024emma,wu2025language,zhu2023difftraj,yang2025trajectory}. Text-based control has also been studied for human motion generation, where language guides body motion synthesis and editing~\cite{jiang2023motiongpt,wan2024tlcontrol,athanasiou2024motionfix,chen2024motionclr}. 
Pedestrian trajectory modeling follows this trend. LMTraj~\cite{bae2024lmtrajectory} reformulates forecasting in language space, LG-Traj~\cite{chib2025lg} incorporates LLM-derived motion cues, and recent methods combine textual instructions with visual scene context~\cite{moon2024visiontrap,shenkut2025visual}. Beyond prediction, Text-Crowd~\cite{ji2024text} and CrowdMoGen~\cite{cao2024crowdmogen} generate pedestrian trajectories from text. These developments motivate evaluating whether generated trajectories faithfully reflect specified textual conditions.
\vspace{-1.4em}
\paragraph{Trajectory datasets and evaluation.}
Established pedestrian datasets~\cite{pellegrini2009you, lerner2007crowds, robicquet2016learning, Kothari2020HumanTF} target predictive accuracy on everyday motion in sidewalks and campuses. Later efforts unify existing datasets~\cite{amirian2020opentraj, ivanovic2023trajdata} but inherit this scope. Event-driven behaviors such as panic egress, violent confrontation, or coordinated demonstration remain largely absent because they are difficult to capture in the wild. Evaluation has followed a similar arc. Per-agent metrics such as ADE and FDE measure geometric proximity to ground truth~\cite{alahi2016social,gupta2018social}. Scene-level realism is assessed through collision rate, KL divergence, kinematic consistency, and recent measures such as density, coverage, and Earth Mover's Distance~\cite{bae2025crowdes,rubner1998metric}, with diversity metrics used to detect mode collapse. These metrics compare generated trajectories against reference distributions, but do not directly test whether trajectories reflect the semantic content of textual prompts. This gap motivates \stride{}.
\vspace{-1.2em}
\paragraph{Pedestrian sociology and pedestrian dynamics.}
Pedestrian trajectories implicitly reflect heterogeneous human behavior,
especially in multi-agent scenarios. Collective human motion has been studied
along two largely separate lines. Sociological accounts, from Canetti's typology
of crowds~\cite{canetti1984crowds} to Le Bon's contagion theory~\cite{lebon1895crowd},
characterize how people gather and behave in groups. McPhail and Wohlstein
further define observable dimensions of gathering behavior, such as direction,
velocity, and temporal change~\cite{McPhailWohlstein1983, mcphail2007crowd},
providing theoretical grounding for our \protocol{} protocol. Pedestrian
dynamics, by contrast, formalizes locomotion through physical and rule-based
simulators, especially Helbing's work on the Social Force Model~\cite{helbing1995social,helbing2005self}.
\stride{} connects these traditions by drawing behavioral categories from sociology
and operationalizing them as trajectory-level measurements, thereby evaluating
aspects of pedestrian behavior that physics-grounded metrics alone may miss.

% \vspace{-1.1em}
\section{\stride{} Framework}

\begin{figure}[t]
  \centering
  \includegraphics[width=\columnwidth]{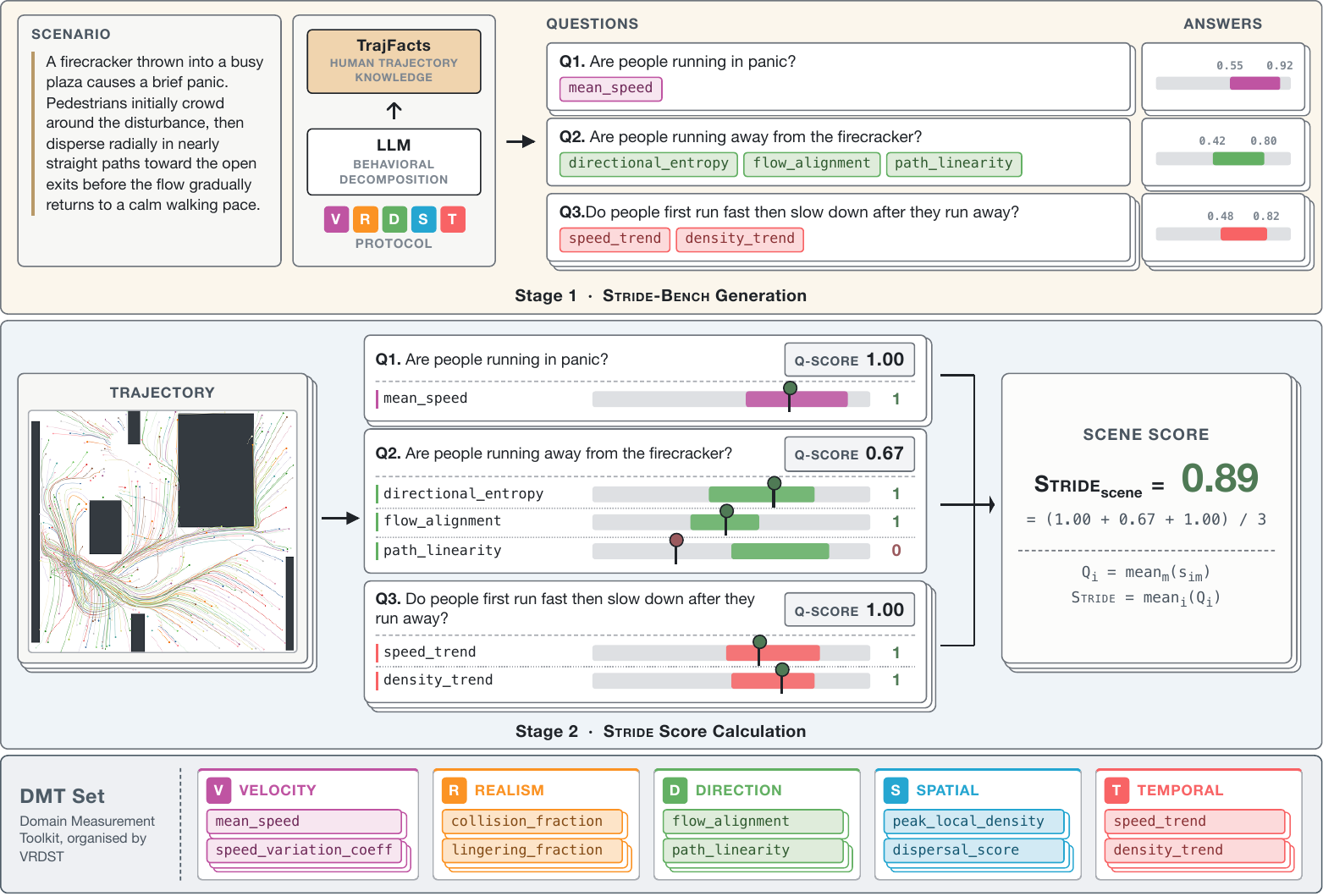}
  \caption{Overview of the \stride{} evaluation framework. Stage 1 constructs \stridebench{} by decomposing each scenario into behavioral questions, measurement functions, and expected answers. Stage 2 computes the \stride{} score by comparing function outputs against the expected answers.}
  \label{fig:STRIDE-framework}
   \vspace{-1.3em}
\end{figure}

\vspace{-0.5em}
We introduce \stride{}, a framework for text-to-trajectory alignment evaluation through behavioral decomposition and structured verification. Figure~\ref{fig:STRIDE-framework} provides an overview.

\vspace{-1em}
\subsection{Design Insights}
\label{sec:insights} 
\vspace{-0.5em}
Evaluating trajectories requires assessing whether sequences of coordinates reflect context-dependent human behavior. Direct LLM-as-judge is insufficient: without scenario-calibrated metric selection, irrelevant or overrepresented attributes can dominate the final judgment. Appendix~\ref{app:preliminary} provides an example. \stride{} therefore formulates alignment evaluation as \textbf{scenario-adaptive and structured behavioral validation}. It addresses three questions: (i) how to define the evaluation dimensions, (ii) how to instantiate it for each scenario, and (iii) how to verify behavior reliably from trajectories.
First, we ground the evaluation space in pedestrian sociology and distill it into a five-axis protocol, \protocol{}. Second, we use this protocol to guide the generation of scenario-curated behavioral questions. Third, each question is verified using our deterministic measurement tool (DMT) library, where each DMT function operates directly on trajectory coordinates and returns an exact numerical statistic. An LLM-generated threshold is then applied to convert the statistic into a binary verdict. This design yields a structured, reproducible, and numerically grounded framework for evaluating text-to-trajectory alignment.

\vspace{-0.45cm}
\subsection[Evaluation Space: VRDST Protocol]{Evaluation Space: \protocol{} Protocol}
\label{sec:defining-metrics}
\vspace{-0.5em}
\textbf{Theoretical Foundation: Pedestrian Sociology.}
To define a verifiable behavioral evaluation space, we derive a protocol from social-scientific theories of crowds and pedestrian dynamics. As its sociological backbone, we adopt McPhail and Wohlstein's theory~\cite{McPhailWohlstein1983}, which identifies \emph{direction}, \emph{velocity}, \emph{time}, and \emph{substantive content} as basic dimensions of gathering behavior, later incorporated into the Elementary Forms of Collective Action (EFCA) framework~\cite{mcphail2007crowd}. We further incorporate pedestrian dynamics theories that capture the computational and physical aspects of motion: realism-related metrics such as collision avoidance and lingering draw on Hall's theory of personal space~\cite{hall1966hidden}; flow- and regime-level metrics such as lane formation and evacuation time draw on studies of normal and evacuation dynamics~\cite{helbing2000simulating, helbing2007dynamics, bandini2019collision, moussaid2010walking}; and spatial-structure metrics such as clustering, group formation, and density patterns are informed by studies of collective pedestrian behavior~\cite{sieben2017collective}. Together, these theories yield a \textbf{social-theory-grounded, layered} protocol for evaluating pedestrian trajectories: realism establishes basic physical validity; velocity and direction capture individual motion; spatial and temporal structure capture how trajectories are organized and evolve within the scene.

\noindent\textbf{\protocolsingle{V}~Velocity.} 
\emph{Do individual pedestrians move at speeds consistent with the scenario's activity regime?}
Velocity captures the scalar component of individual motion. A campus afternoon
implies free-flow walking in the 1.10 to 1.65~m/s range~\cite{pellegrini2009you},
whereas an explosion in a transit hub implies a higher mean speed with a heavy
right tail. Velocity is the most directly measurable dimension in McPhail and
Wohlstein's original taxonomy~\cite{McPhailWohlstein1983}.

\noindent\textbf{\protocolsingle{R}~Realism.} 
\emph{Do pedestrians behave in physically feasible ways within the environment?}
Realism captures whether pedestrian behavior respects scene geometry, obstacles,
and motion constraints, while avoiding artifacts such as overlap, wall penetration,
teleportation, or impossible accelerations. Because feasible behavior depends on
environmental structure, this axis is grounded in proxemics~\cite{hall1966hidden}
and evacuation dynamics~\cite{helbing2000simulating}.

\noindent\textbf{\protocolsingle{D}~Direction.} 
\emph{Do individual pedestrians move in directions implied by the scenario?}
Direction captures the vector component of individual motion. Unstructured
scenarios allow heterogeneous headings, while directed scenarios, such as
evacuations, require coordinated movement toward exits or other scenario-specific
goals. This axis follows McPhail and Wohlstein's direction dimension~\cite{McPhailWohlstein1983}
and flow-alignment measures from self-organized pedestrian dynamics~\cite{helbing2005self}.

\noindent\textbf{\protocolsingle{S}~Spatial.} 
\emph{Are pedestrians distributed according to the scenario's spatial context?}
Spatial structure evaluates whether generated trajectories occupy plausible
regions and form the expected spatial pattern. For example, evacuation trajectories
should move away from hazards toward exits, gathering scenarios should place
pedestrians near points of interest, and leisure settings may allow dispersed
movement. This axis draws on spatial structure and group-formation studies~\cite{sieben2017collective,moussaid2010walking}.

\noindent\textbf{\protocolsingle{T}~Temporal.} 
\emph{Do generated trajectories evolve over time in a scenario-consistent way?}
Temporal structure evaluates whether the above properties change coherently as
the scenario unfolds. EFCA treats collective descriptors as time-dependent~\cite{mcphail2007crowd}.
We operationalize this axis by tracking trends in realism, velocity, direction,
and spatial structure over sliding windows, enabling consistent evaluation across
both short-horizon prediction and long-horizon trajectory generation models.

\vspace{-1em}
\subsection{Scenario-Adaptive Behavioural Decomposation}
\label{sec:scenario-curated-questions}
\vspace{-0.5em}
As discussed in Section~\ref{sec:insights}, \stride{} does not weigh all dimensions
equally across scenarios. Pedestrian behavior is highly context-dependent: speed
may be critical in an evacuation, group cohesion in a guided tour, and lane
formation in bidirectional flow. Appendix~\ref{app:preliminary} shows that equal
weighting across dimensions can be unstable and biased, motivating \stride{}'s scenario-adaptive evaluation criteria.
For each scenario, we prompt an LLM with the scenario description and the
\protocol{} protocol to generate scenario-specific evaluation questions. The
model returns at least five questions following the protocol and justifies each
decomposition for traceability. These questions describe expected behavioral
properties in semantic terms, for example: \emph{``Do pedestrians form denser
groups near the projection area while following plausible paths around
obstacles?''}

\vspace{-1em}
\subsection{Verification Space: Deterministic Measurements}
\label{sec:how-to-eval}
\vspace{-0.5em}
% We first motivate the static verification
% design, then present the function
% library (DMT) that realizes the five axes, and
% finally define how to calculate the \stride{} score.

The \protocol{} protocol specifies \emph{what} perspectives to evaluate; this section specifies \emph{how} to evaluate them reliably.
\textbf{Why deterministic verification functions?}
Benchmark scores are meaningful only if they are trustworthy and
comparable. \stride{} grounds each per-question verdict in a deterministic function
of trajectory coordinates, yielding three benefits. First, \emph{reproducibility}:
a fixed trajectory-question pair always receives the same score, independent of
prompts, evaluation runs, or judge-model versions. Second, \emph{numerical
faithfulness}: quantities such as mean speed, collision count, and flow alignment
are computed directly from coordinates, making each verdict traceable to the
value that triggered it. Third, \emph{scale invariance}: the same functions
operate across different time horizons and agent counts, allowing short
prediction windows and long generation rollouts to be evaluated on a common
basis.
\textbf{Deterministic Measurement Tool Library (DMT).}
We implement 20 deterministic measurement functions, each operating on trajectory
coordinates and assigned to one protocol axis. The resulting DMT library
operationalizes \protocol{} as reusable measurement primitives. DMT is
compositional: for example, ``people rush away from the hazard'' can be evaluated
through elevated speed, outward flow relative to the hazard, and a centrifugal
density pattern, each measured independently and then combined in the \stride{}
score. Thus, DMT covers diverse behaviors by composing reusable measurements
with scenario-specific expectations.
\textbf{Verifiable Answer Space.}
Given the DMT library, \stride{} defines a verifiable answer space for the
decomposed behavioral questions. For each question, a frontier LLM receives the
scenario description, the question, and the DMT specifications, including each
function's semantics, output type, and applicable protocol axis. It also receives
\emph{TrajFacts}. This human-trajectory knowledge base stores reference values used to calibrate expectations, such as typical walking speeds, plausible collision
rates, and scenario-dependent density patterns. See Appendix~\ref{app:gt-library}
for details. The LLM then selects one or more relevant DMT functions and converts
the semantic expectation into numerical answer ranges over their outputs, with a
justification for traceability.
\textbf{\stride{} Score.}
With these measurement functions, \stride{} scores each question by comparing its
expected answer ranges with the computed trajectory values. The \stride{} score is
aggregated hierarchically: measurements are averaged within each question,
questions within each scenario, and scenarios across the benchmark.
Appendix~\ref{app:trajQA-score} gives the mathematical formulation.

\begin{figure*}[t]
  \centering
  \includegraphics[width=\columnwidth]{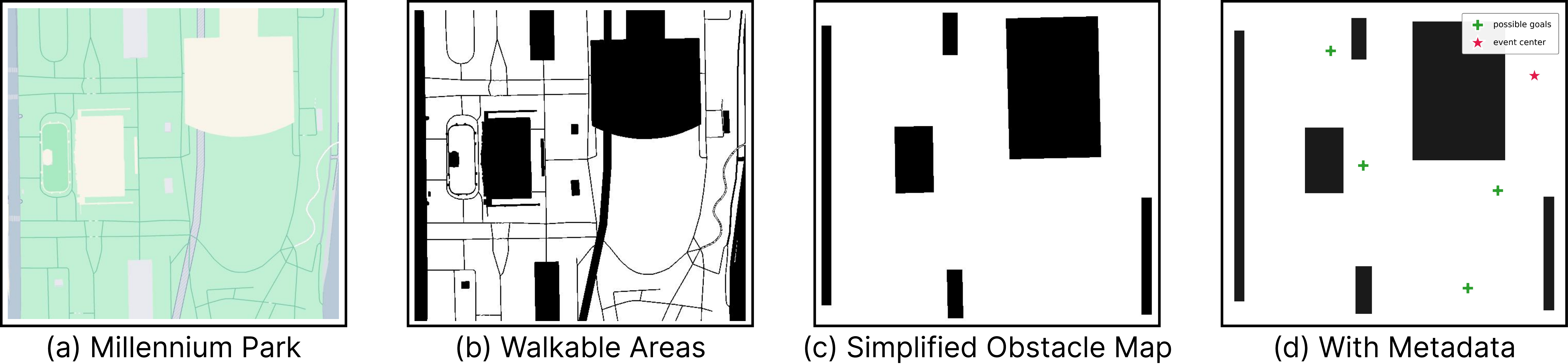}
  \caption{Scenario corpus map processing: (a) real-world map from Google Map. (b) walkable areas defined by Google Map color set. (c) polygon-approximated obstacles. (d) metadata on the map.}
  \label{fig:real-world-obstacle}
  \vspace{-1.5em}
\end{figure*}

\vspace{-1em}
\subsection{Benchmark Construction and Usage}
\label{sec:benchmark-construction}
\begin{figure}[t]
  \centering
  \includegraphics[width=\columnwidth]{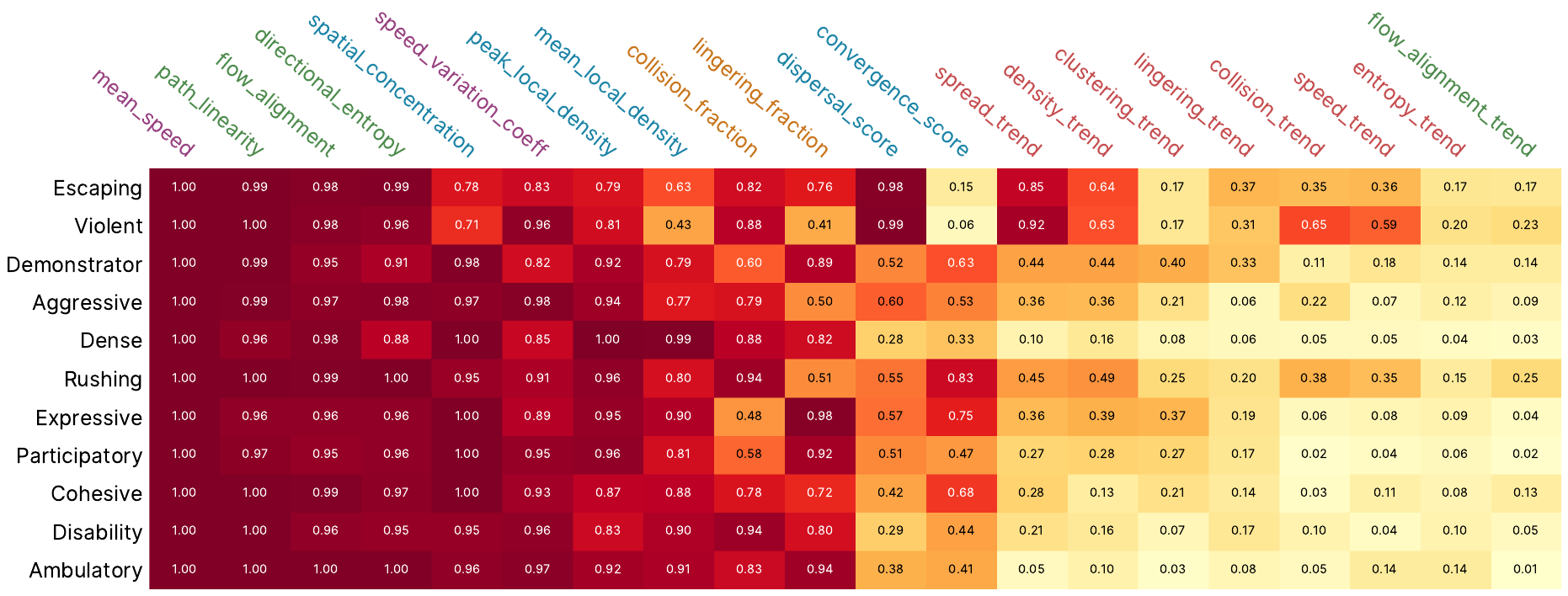}
  \caption{DMT function usage across crowd categories in \stridebench{}. Each cell reports the percentage of scenes within a category that use the corresponding function.}
  \label{fig:benchmark-stats}
  \vspace{-1.5em}
\end{figure}

\vspace{-0.5em}
We instantiate \stride{} as a benchmark for pedestrian trajectory generation in
crowd scenarios. We first construct a corpus of crowd scenario descriptions,
each paired with a map and metadata such as event center and potential goals. For each scenario, we then generate
evaluation questions, select measurements, and derive expected answers calibrated
with real-world crowd references.
\textbf{Maps.}
Beyond text descriptions, trajectory generation models typically require map inputs to ground the generated trajectories in physical space. Thus, we curate 113 maps of high-traffic public venues, including major tourist
destinations, stadiums, university campuses, and transit hubs. These venues
routinely host large crowds during events. The maps are acquired via the Google Maps API with a fixed size of
$301.7~\text{m} \times 282.8~\text{m}$, and converted into obstacle
representations using a three-stage pipeline: semantic segmentation based on the
Google Maps color scheme, polygon approximation of segmented obstacle regions,
and rasterization into obstacle masks. Figure~\ref{fig:real-world-obstacle}
illustrates this map-processing pipeline.
\uline{\textbf{Scenario descriptions.}}
We adopt Berlonghi's typology of 11 crowd categories~\cite{berlonghi1995understanding}. For each category, we prompt GPT-5.1 with a structured template that elicits: (i) a natural-language scenario description, (ii) candidate exits and event-center coordinates in the map pixel grid, and (iii) the initial crowd size and spatial distribution. Fields (ii) and (iii) provide a common initialization across models, allowing us to isolate context-conditioned behavior from setup variability. 
\textbf{Filtering and stratified sampling.}
We filter out scenarios with insufficient pedestrian density, exits or event
centers that fall on obstacles, and near-duplicate descriptions. Obstacle
conflicts can arise from polygon approximation errors, while near-duplicates are
identified by cosine similarity over sentence embeddings. We also conduct a
manual pass to remove remaining cases with implausible map annotations.
From the filtered pool, we sample more heavily from safety-critical crowd
categories, such as Escaping, Violent, and Dense. These categories are
underrepresented in real-world trajectory datasets~\cite{pellegrini2009you,
lerner2007crowds, robicquet2016learning}, since such dynamics cannot be
ethically staged. They are also important for applications such as urban
planning, venue safety auditing, and emergency preparedness. For each retained
scenario, we prompt GPT-5.2 to generate the behavioral questions, measurements,
and expected answers.
\uline{\textbf{Calibration.}}
To align expected answers with real-world crowd behavior, we incorporate
crowd-reference data into our human-trajectory knowledge base, \emph{TrajFacts}.
These references include reported measurements from events such as the Love
Parade~\cite{zhao2020assessing, krausz2012loveparade} and the Itaewon crowd
crush~\cite{hwang2025virtual}. \emph{TrajFacts} provides reference values used to
calibrate expected-answer ranges. Appendix~\ref{app:gt-library} gives additional
details.
In total, \stridebench{} contains 936 scenarios, 6,633 questions, and 11,696 measurements. Figure~\ref{fig:benchmark-stats} shows the heatmap of DMT function usage across crowd categories.
\uline{\textbf{Usage.}}
\stride{} supports two evaluation modes. In \emph{benchmark mode}, evaluation uses the released scenarios, cached questions, and expected-answer ranges. The final score is therefore produced entirely by deterministic function calls, with no LLM required at scoring time. In \emph{open mode}, an LLM agent generates the question set on the fly from a user-supplied scenario description, enabling evaluation beyond the released benchmark.

% \begin{wrapfigure}[13]{r}{0.5\textwidth}
%     \vspace{-2em}
%     \includegraphics[width=0.5\textwidth]{figures/fig_bubble_questions_measurements.pdf}
%     \caption{The bubble plot shows the count of question and measurements for each scene}
%     \label{fig:bubble}
% \end{wrapfigure}
\vspace{-1em}
\section{Validation}
\label{sec: validation}
\vspace{-1em}
Given the diversity and context dependence of pedestrian behavior, alignment metrics should be validated against human judgment. We validate \stridebench{} in three stages: real-world trajectory validation (Section~\ref{sec:human-traj-validation}),
human ratings with inter-annotator and benchmark agreement
(Section~\ref{subsec:human-validation}), and judge-model robustness across
state-of-the-art LLMs (Section~\ref{subsec:llm-agreement}).

\vspace{-1em}
\subsection{Real-world Human Trajectory Validation}
\label{sec:human-traj-validation}
\vspace{-0.5em}

To assess whether \stridebench{} assigns high alignment scores to real crowd
behavior, we evaluate it on trajectories from the \emph{Fête des Lumières in
Lyon}, a large public gathering documented from three camera views~\cite{dufour_2024_dense_crowd_dynamics} as well as the ETH/UCY datasets~\cite{lerner2007crowds, pellegrini2009you}, which consist of bird's-eye-view video recordings of pedestrians in everyday public spaces under ordinary, non-emergency conditions.

\textbf{Fête des Lumières in Lyon} The dataset contains 12 trajectory recordings. For each recording, we construct a scenario description and apply the same QA generation
procedure as in Section~\ref{sec:benchmark-construction}, yielding 86 questions
and 154 measurements.
Real human trajectories achieve an average \stride{} score of $0.94$ across the 12
recordings, with per-recording scores ranging from $0.87$ to $1.00$. This
suggests that \stridebench{} assigns high alignment scores to real crowd behavior.
Figure~\ref{fig:traj-comparison}(a) shows real human
trajectories from \emph{Large View Tracers}. Additional details are provided in Appendix~\ref{app:lyon-validation}. \textbf{ETH/UCY} The dataset consists of five standard scenes (eth, hotel, univ, zara1, zara2) across four venues in Switzerland and Cyprus. We built a benchmark from the dataset's scenarios and used it to score the real human trajectories. All its measurements are generated in open mode, without any calibration. Table~\ref{tab:ethucy-stride} in Appendix~\ref{app:ethucy} shows the \stride{} score for all scenarios. ETH/UCY averages $0.914$, close to Lyon's $0.943$, proving the consistency of the \stride{} score on human ground truth.
A perfect score ($1.0$) is not expected, since expected ranges are inferred from textual descriptions and calibrated reference values, which may not capture all
scene-specific statistics.

\vspace{-0.8em} 
\subsection{Human Validation}
\label{subsec:human-validation}
\vspace{-0.5em}
To validate the fidelity of the LLM-generated questions and expected answers, we conducted a human annotation study on \href{https://www.prolific.com}{Prolific}.
We recruited $n=50$ annotators through the platform. Each annotator was
compensated at \$14/hour and completed at least three \stridebench{} scenarios. For each scenario, annotators rated task-specific reference information, such as typical walking speed. We selected scenarios annotated by at least three independent annotators, yielding 15{,}750 annotations across 56 scenarios and 350 questions.
Before the main study, we conducted a pilot study to refine the annotation
protocol and improve task interpretability since the benchmark contains
mathematical terminology and numerical expected answers that are difficult for
humans to interpret directly. Based on the pilot, we designed a hybrid interface
that combines natural-language descriptions with task-specific visualizations.
The full annotation interface is provided in Appendix~\ref{app:interface-design}.
Figure~\ref{fig:human-annotation-results}(a) summarizes the results. The left
panel reports agreement between human annotators and \stridebench{} answers,
together with Cohen's $\kappa$, which measures agreement beyond chance. The
right panel reports inter-annotator agreement, together with Krippendorff's
$\alpha$, which measures reliability among multiple annotators.
\textbf{Inter-annotator agreement.}
We first measure consistency among independent annotators. Pairwise
inter-annotator agreement reaches 66\%, with Krippendorff's $\alpha=0.698$.
Since $\alpha$ accounts for disagreement across multiple annotators, this
indicates reliable agreement despite the inherent ambiguity of fine-grained
crowd behavior judgments.
\textbf{Agreement with benchmark answers.}
We then compare each annotator's response with the corresponding \stridebench{}
answer. Annotators agree with \stridebench{} on 80\% of question-answer pairs,
with Cohen's $\kappa=0.730$. Since $\kappa$ corrects for chance agreement, this
indicates substantial alignment between \stridebench{} answers and human judgments.
Overall, independent annotators agree reliably with one another and show even stronger agreement with \stridebench{} answers. Table~\ref{tab:crowd-category-agreement} reports inter-annotator agreement for each crowd category. Agreement is lowest for \textit{Expressive} ($0.661$), which, owing to the composition of the dataset, has only 62 judgments compared with 256-390 for the other categories, making its estimate the least precise. Nevertheless, annotators still agree on about two thirds of its judgments ($\kappa = 0.522$), and all remaining categories fall between $0.708$ and $0.869$. This suggests that \stridebench{} is consistent with human consensus and provides a reliable reference for behavior-level alignment evaluation.

\begin{table*}[!ht]
\centering
\caption{Per-crowd-category inter-annotator agreement on the benchmark judgments. For each category we report the number of judgments, the agreement rate, Krippendorff's $\alpha$ and cohen's $\kappa$. Category names are abbreviated by their first three letters.}
\label{tab:crowd-category-agreement}
\setlength{\tabcolsep}{3pt}

\resizebox{\textwidth}{!}{
\begin{tabular}{lcccccccccccc}
\toprule
Metric & {Agg.} & {Amb.} & {Coh.} & {Dem.} & {Den.} & {Dis.} & {Esc.} & {Exp.} & {Par.} & {Rus.} & {Vio.} & Overall \\
\midrule
Judgments &
$295$ & $319$ & $268$ & $260$ & $256$ & $314$ & $390$ & $62$ & $312$ & $325$ & $293$ &
$3{,}094$ \\

Agreement &
$0.722$ & $0.784$ & $\textbf{0.869}$ & $0.723$ & $0.824$ & $0.796$ & $0.746$ & $0.661$ & $0.708$ & $0.778$ & $0.819$ &
$0.766 \pm 0.058$ \\

$\alpha$ &
$0.695$ & $0.628$ & $\textbf{0.761}$ & $0.675$ & $0.663$ & $0.682$ & $0.678$ & $0.808$ & $0.740$ & $0.689$ & $0.698$ &
$0.702 \pm 0.048$ \\

$\kappa$ &
$0.703$ & $0.716$ & $\textbf{0.848}$ & $0.617$ & $0.748$ & $0.722$ & $0.675$ & $0.522$ & $0.640$ & $0.734$ & $0.674$ &
$0.691 \pm 0.079$ \\
\bottomrule
\end{tabular}
}
\vspace{-0.3em}
\end{table*}

\begin{figure*}[t]
  \centering
  \includegraphics[width=1\textwidth]{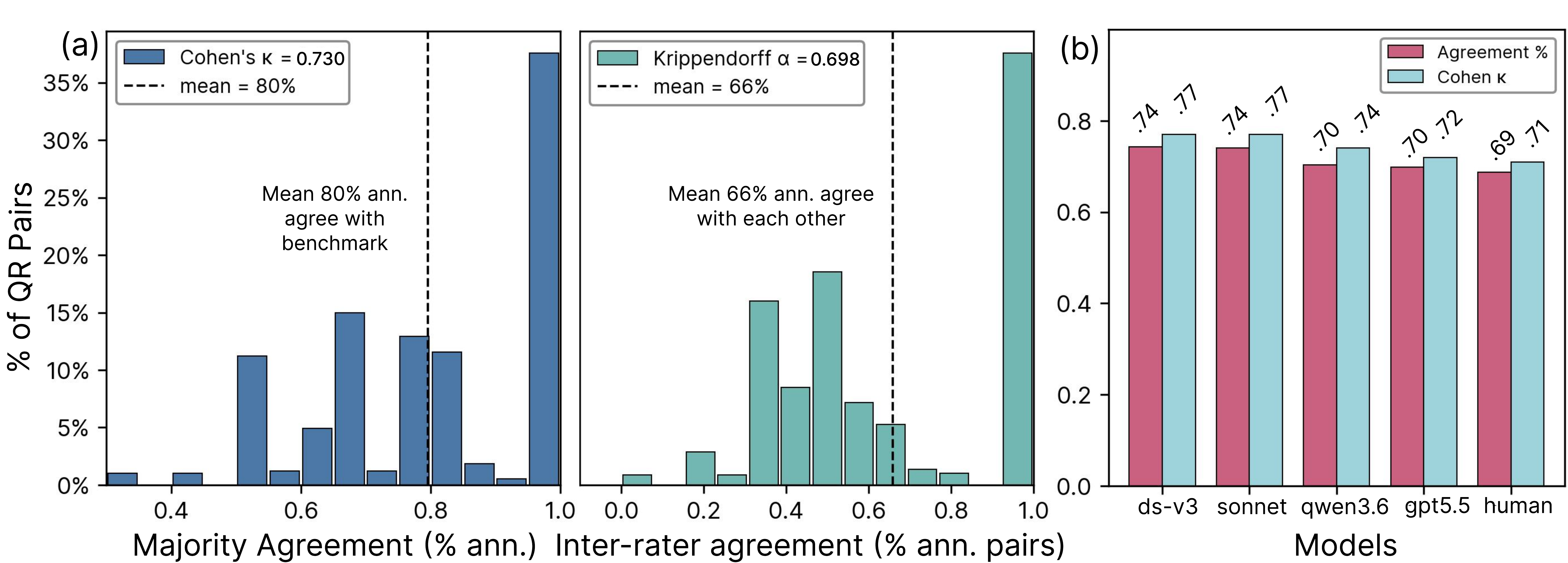}
    \caption{
        \stridebench{} validation (ann. means annotators). 
        (a) Human validation. Annotator-benchmark agreement averages 80\% with Cohen's $\kappa=0.730$; inter-annotator agreement of 66\% with Krippendorff's $\alpha=0.698$, indicating \stridebench{} answers are human-consistent.
        (b) deepseek-v3, claude-sonnet-4.6, qwen3.6-plus, and gpt-5.5 achieve similar agreement and Cohen's $\kappa$ with the benchmark, comparable to human annotators, showing robustness across models.
    }
  \label{fig:human-annotation-results}
  \vspace{-1em}
\end{figure*}

\vspace{-0.8em}
\subsection{Cross-LLM Robustness}
\label{subsec:llm-agreement}
\vspace{-0.5em}
To assess the cross-LLM robustness of \stridebench{}, we examine whether its reliability generalizes beyond GPT-5.2. Specifically, we evaluate four state-of-the-art LLMs using the same questionnaire as in Section~\ref{subsec:human-validation}, measuring their agreement with the consensus human label on $n{=}588$ items. As a human reference, we compute a leave-one-out baseline, where each annotator is compared against the majority vote of the remaining annotators. We report both raw agreement and Cohen's $\kappa$.
Figure~\ref{fig:human-annotation-results}(b) shows that all four LLMs fall within the leave-one-out human range on both metrics, with tightly clustered scores across models. This indicates that the evaluated LLMs achieve human-comparable agreement on \stride{} items and produce stable judgments across model families. Together with the human-benchmark agreement in Section~\ref{subsec:human-validation}, these results suggest that \stride{}'s reliability is not tied to a single generation or judge model, supporting its \emph{open mode} for user-specified scenarios. This further suggests that \stride{} can support extensible evaluation when new scenarios are generated with different LLM backends.

\vspace{-0.5em}
\section{Evaluation}
\vspace{-0.3cm}

\begin{figure*}[!ht]
    \centering
    \includegraphics[width=1\columnwidth]{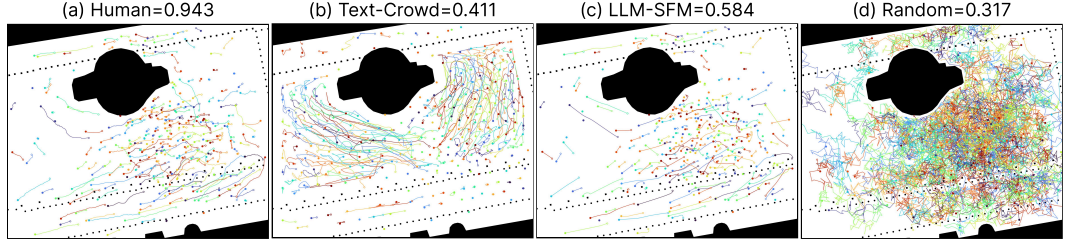}
    \caption{Trajectories for \emph{Fête des Lumières in Lyon; Large View Trackers}. \stride{} scores are (a) Human, $0.943$; (b) Text-Crowd, $0.411$; (c) LLM-SFM, $0.584$; (d) Random Walk, $0.317$.}
    \label{fig:traj-comparison}
    \vspace{-0.5em}
\end{figure*}

% We use \stridebench{} to evaluate five trajectory generation models and analyze their context alignment across multiple behavioral dimensions.
\begin{table*}[!ht]
\centering
\caption{Per-category and overall STRIDE-Bench scores, where higher is better ($\mathrm{STRIDE}_{\mathrm{score}}\in[0,1]$). Category names are abbreviated by their first three letters. The last column reports overall mean $\pm$ standard deviation on the benchmark dataset.} 
\label{tab:crowd-category-x-overall}
\setlength{\tabcolsep}{3pt}

\resizebox{\textwidth}{!}{
\begin{tabular}{lcccccccccccc}
\toprule
Model & {Agg.} & {Amb.} & {Coh.} & {Dem.} & {Den.} & {Dis.} & {Esc.} & {Exp.} & {Par.} & {Rus.} & {Vio.} & $\mathrm{STRIDE}_{\mathrm{score}}\uparrow$ \\
\midrule
Text-Crowd &
$\mathbf{0.669}$ & $\mathbf{0.787}$ & $\mathbf{0.733}$ & $\mathbf{0.771}$ & $\mathbf{0.729}$ &
$\mathbf{0.778}$ & $\mathbf{0.393}$ & $\mathbf{0.829}$ & $\mathbf{0.790}$ & $\mathbf{0.584}$ & $\mathbf{0.387}$ &
$\mathbf{0.645 \pm 0.233}$ \\

LLM-SFM &
$0.488$ & $0.327$ & $0.418$ & $0.605$ & $0.650$ &
$0.489$ & $0.256$ & $0.610$ & $0.568$ & $0.243$ & $0.230$ &
$0.434 \pm 0.231$ \\

SingularTrajectory &
$0.425$ & $0.452$ & $0.526$ & $0.540$ & $0.530$ &
$0.505$ & $0.238$ & $0.536$ & $0.509$ & $0.358$ & $0.229$ &
$0.421 \pm 0.177$ \\

Random Walk &
$0.355$ & $0.409$ & $0.348$ & $0.312$ & $0.273$ &
$0.368$ & $0.268$ & $0.340$ & $0.312$ & $0.312$ & $0.254$ &
$0.311 \pm 0.117$ \\

Stop &
$0.259$ & $0.260$ & $0.249$ & $0.230$ & $0.194$ &
$0.307$ & $0.178$ & $0.237$ & $0.246$ & $0.195$ & $0.185$ &
$0.221 \pm 0.105$ \\
\bottomrule
\end{tabular}
}
\vspace{-0.3em}
\end{table*} 

\vspace{-0.2cm} 
\subsection{Experiment Setup}
\vspace{-0.1cm}
\label{sec: baseline}
LLM-SFM. We use an LLM to map each scenario to per-agent parameters of the Social Force Model (SFM)~\cite{helbing1995social}, then simulate trajectories with an SFM simulator. Text-Crowd~\cite{ji2024text} generates group-level distributions and velocity fields from maps and text via LLM canonicalization and a conditional diffusion model. 
SingularTrajectory~\cite{bae2024singulartrajectory} is a diffusion-based universal trajectory predictor. We use it to test whether trajectory history alone contains implicit behavioral information sufficient to recover scenario-consistent behavior without text input.
Statistical baselines.
Random Walk samples random speeds and headings at each step and stops agents at obstacles. Stop keeps all agents fixed.

\vspace{-0.2cm}
\subsection{Results and Analysis}
\vspace{-0.2cm}
\label{sec:results}

We evaluate these baselines on \stridebench{} using overall and per-category scores (Table~\ref{tab:crowd-category-x-overall}), rollout horizon (Table~\ref{tab:time_x_vrdst}), and crowd size (Table~\ref{tab:agent_x_vrdst}). The horizon and scale partitions follow standard crowd-dynamics regimes~\cite{helbing2000simulating,helbing2005self,helbing2007dynamics}. 
\textbf{Overall and per-category performance.} Text-Crowd achieves the highest score across all crowd categories, but its overall \stride{} score remains $0.645$, indicating that current text-to-trajectory models still fall substantially short of reliable context-aligned generation. SingularTrajectory performs competitively with LLM-SFM despite not using text input, suggesting that trajectory history alone captures part of the scenario context. 
\textbf{Effect of rollout horizon.} Table~\ref{tab:time_x_vrdst} shows that model rankings vary with rollout length. LLM-SFM performs best in short-to-medium horizons, while Text-Crowd is stronger at long horizons, suggesting better preservation of high-level intent over extended rollouts. 
\textbf{Effect of crowd size.} Table~\ref{tab:agent_x_vrdst} shows that Text-Crowd performs best across all crowd-size regimes. LLM-SFM improves most in mass-gathering settings, consistent with social force models, while SingularTrajectory also benefits from larger crowds, suggesting that aggregate motion patterns provide useful context. 
\textbf{Fête des Lumières in Lyon.} We further evaluate the baselines on the 12 \emph{Fête des Lumières in Lyon} scenes from Section~\ref{sec:human-traj-validation}. The average \stride{} scores are: Human $0.943$, LLM-SFM $0.584$, Text-Crowd $0.411$, and Random Walk $0.317$. The gap between human and generated trajectories reinforces the limitations of existing models. In this setting, Text-Crowd underperforms LLM-SFM, likely due to sensitivity to complex map layouts. Figure~\ref{fig:traj-comparison} shows both human and model-generated trajectories.
\textbf{Main findings.} Overall, recent trajectory generation models exhibit nascent but limited capability in context-aligned generation. The strongest baseline reaches only $0.645$ on \stridebench{} and remains far below real human trajectories. Fine-grained \stride{} scores further reveal persistent failures in complex behavioral requirements, including spatial reasoning and fine-grained context conditioning. More detailed results can be found in Appendix~\ref{app:eval-results}, including the DMT results differences across models.

\begin{table*}[t]
\centering
\caption{STRIDE scores across models, time windows, and \protocol{} axes. Values are mean $\pm$ standard deviation; higher is better. Mean denotes the average score among the protocols.}
\label{tab:time_x_vrdst}
\small
\resizebox{\textwidth}{!}{
\begin{tabular}{llcccccc}
\toprule
Time & Model
& \protocolsingle{V}
& \protocolsingle{R}
& \protocolsingle{D}
& \protocolsingle{S}
& \protocolsingle{T}
& Mean$\uparrow$ \\
\midrule

\multirow{5}{*}{0--30s}
 & LLM-SFM       & $\mathbf{0.615} \pm 0.360$ & $\mathbf{0.860} \pm 0.083$ & $0.523 \pm 0.265$ & $0.603 \pm 0.136$ & $0.268 \pm 0.191$ & $\mathbf{0.574} \pm 0.190$ \\
 & SingularTraj  & $0.217 \pm 0.203$ & $0.819 \pm 0.123$ & $0.379 \pm 0.197$ & $\mathbf{0.604} \pm 0.131$ & $0.251 \pm 0.211$ & $0.454 \pm 0.228$ \\
 & Text-Crowd    & $0.531 \pm 0.070$ & $0.734 \pm 0.078$ & $\mathbf{0.677} \pm 0.182$ & $0.507 \pm 0.049$ & $\mathbf{0.288} \pm 0.157$ & $0.547 \pm 0.156$ \\
 & Random        & $0.504 \pm 0.480$ & $0.817 \pm 0.126$ & $0.019 \pm 0.021$ & $0.376 \pm 0.141$ & $0.093 \pm 0.079$ & $0.362 \pm 0.289$ \\
 & Stop          & $0.012 \pm 0.009$ & $0.471 \pm 0.471$ & $0.182 \pm 0.233$ & $0.350 \pm 0.154$ & $0.026 \pm 0.041$ & $0.208 \pm 0.180$ \\

\midrule

\multirow{5}{*}{30--120s}
 & LLM-SFM       & $\mathbf{0.806} \pm 0.124$ & $\mathbf{0.914} \pm 0.027$ & $0.471 \pm 0.203$ & $0.687 \pm 0.095$ & $0.283 \pm 0.152$ & $\mathbf{0.632} \pm 0.228$ \\
 & SingularTraj  & $0.297 \pm 0.289$ & $0.825 \pm 0.118$ & $0.249 \pm 0.087$ & $\mathbf{0.690} \pm 0.089$ & $0.257 \pm 0.162$ & $0.464 \pm 0.244$ \\
 & Text-Crowd    & $0.696 \pm 0.051$ & $0.829 \pm 0.091$ & $\mathbf{0.604} \pm 0.152$ & $0.606 \pm 0.098$ & $\mathbf{0.360} \pm 0.147$ & $0.619 \pm 0.153$ \\
 & Random        & $0.504 \pm 0.480$ & $0.817 \pm 0.126$ & $0.020 \pm 0.021$ & $0.419 \pm 0.148$ & $0.131 \pm 0.116$ & $0.378 \pm 0.283$ \\
 & Stop          & $0.012 \pm 0.009$ & $0.471 \pm 0.471$ & $0.182 \pm 0.233$ & $0.350 \pm 0.154$ & $0.026 \pm 0.041$ & $0.208 \pm 0.180$ \\

\midrule

\multirow{5}{*}{120s+}
 & LLM-SFM       & $\mathbf{0.534} \pm 0.020$ & $0.695 \pm 0.021$ & $0.146 \pm 0.051$ & $0.596 \pm 0.084$ & $0.295 \pm 0.197$ & $0.453 \pm 0.202$ \\
 & SingularTraj  & $0.373 \pm 0.371$ & $0.786 \pm 0.016$ & $0.070 \pm 0.056$ & $\mathbf{0.611} \pm 0.083$ & $0.282 \pm 0.187$ & $0.424 \pm 0.251$ \\
 & Text-Crowd    & $0.502 \pm 0.114$ & $\mathbf{0.872} \pm 0.031$ & $\mathbf{0.343} \pm 0.023$ & $0.292 \pm 0.151$ & $\mathbf{0.308} \pm 0.250$ & $\mathbf{0.463} \pm 0.217$ \\
 & Random        & $0.504 \pm 0.480$ & $0.817 \pm 0.126$ & $0.020 \pm 0.022$ & $0.444 \pm 0.171$ & $0.153 \pm 0.131$ & $0.388 \pm 0.280$ \\
 & Stop          & $0.012 \pm 0.009$ & $0.471 \pm 0.471$ & $0.182 \pm 0.233$ & $0.350 \pm 0.154$ & $0.026 \pm 0.041$ & $0.208 \pm 0.180$ \\

\bottomrule
\end{tabular}
}
\vspace{-1em}
\end{table*}
\begin{table*}[t]
\centering
\caption{STRIDE scores across models, agent sizes, and \protocol{} axes. Values are mean $\pm$ standard deviation; higher is better. Mean denotes the average score among the protocols.}
\label{tab:agent_x_vrdst}
\small
\setlength{\tabcolsep}{3pt}
\renewcommand{\arraystretch}{1.08}
\resizebox{\textwidth}{!}{
\begin{tabular}{llcccccc}
\toprule
Size & Model
& \protocolsingle{V}
& \protocolsingle{R}
& \protocolsingle{D}
& \protocolsingle{S}
& \protocolsingle{T}
& Mean$\uparrow$ \\
\midrule

\multirow{5}{*}{\begin{tabular}{@{}c@{}}Small \\ (1--25)\end{tabular}}
 & LLM-SFM       & $0.458 \pm 0.085$ & $0.671 \pm 0.023$ & $0.323 \pm 0.094$ & $0.491 \pm 0.195$ & $0.280 \pm 0.301$ & $0.445 \pm 0.138$ \\
 & SingularTraj  & $0.365 \pm 0.365$ & $0.736 \pm 0.042$ & $0.139 \pm 0.123$ & $0.492 \pm 0.196$ & $0.129 \pm 0.083$ & $0.372 \pm 0.228$ \\
 & Text-Crowd    & $\mathbf{0.734} \pm 0.013$ & $\mathbf{0.915} \pm 0.072$ & $\mathbf{0.632} \pm 0.068$ & $\mathbf{0.546} \pm 0.204$ & $\mathbf{0.362} \pm 0.271$ & $\mathbf{0.638} \pm 0.185$ \\
 & Random        & $0.513 \pm 0.487$ & $0.822 \pm 0.165$ & $0.064 \pm 0.075$ & $0.519 \pm 0.139$ & $0.269 \pm 0.329$ & $0.437 \pm 0.256$ \\
 & Stop          & $0.018 \pm 0.018$ & $0.493 \pm 0.493$ & $0.191 \pm 0.214$ & $0.531 \pm 0.261$ & $0.165 \pm 0.345$ & $0.280 \pm 0.199$ \\

\midrule

\multirow{5}{*}{\begin{tabular}{@{}c@{}}Medium \\ (26--100)\end{tabular}}
 & LLM-SFM       & $0.483 \pm 0.029$ & $0.671 \pm 0.031$ & $0.142 \pm 0.057$ & $0.584 \pm 0.092$ & $0.321 \pm 0.225$ & $0.440 \pm 0.189$ \\
 & SingularTraj  & $0.356 \pm 0.354$ & $0.801 \pm 0.023$ & $0.068 \pm 0.058$ & $0.600 \pm 0.097$ & $0.315 \pm 0.228$ & $0.428 \pm 0.252$ \\
 & Text-Crowd    & $\mathbf{0.762} \pm 0.001$ & $\mathbf{0.917} \pm 0.026$ & $\mathbf{0.524} \pm 0.171$ & $\mathbf{0.633} \pm 0.121$ & $\mathbf{0.363} \pm 0.136$ & $\mathbf{0.640} \pm 0.191$ \\
 & Random        & $0.507 \pm 0.483$ & $0.826 \pm 0.117$ & $0.016 \pm 0.020$ & $0.447 \pm 0.136$ & $0.155 \pm 0.132$ & $0.390 \pm 0.284$ \\
 & Stop          & $0.012 \pm 0.005$ & $0.471 \pm 0.471$ & $0.179 \pm 0.228$ & $0.387 \pm 0.199$ & $0.019 \pm 0.040$ & $0.214 \pm 0.188$ \\

\midrule

\multirow{5}{*}{\begin{tabular}{@{}c@{}}Large \\ (101--300)\end{tabular}}
 & LLM-SFM       & $0.605 \pm 0.027$ & $0.722 \pm 0.017$ & $0.092 \pm 0.035$ & $0.606 \pm 0.113$ & $0.240 \pm 0.167$ & $0.453 \pm 0.243$ \\
 & SingularTraj  & $0.390 \pm 0.386$ & $0.786 \pm 0.034$ & $0.053 \pm 0.036$ & $\mathbf{0.612} \pm 0.121$ & $0.243 \pm 0.165$ & $0.417 \pm 0.260$ \\
 & Text-Crowd    & $\mathbf{0.776} \pm 0.034$ & $\mathbf{0.891} \pm 0.026$ & $\mathbf{0.463} \pm 0.257$ & $0.601 \pm 0.148$ & $\mathbf{0.459} \pm 0.268$ & $\mathbf{0.638} \pm 0.172$ \\
 & Random        & $0.496 \pm 0.468$ & $0.824 \pm 0.093$ & $0.013 \pm 0.009$ & $0.424 \pm 0.240$ & $0.132 \pm 0.119$ & $0.378 \pm 0.286$ \\
 & Stop          & $0.009 \pm 0.009$ & $0.459 \pm 0.459$ & $0.195 \pm 0.260$ & $0.249 \pm 0.072$ & $0.014 \pm 0.032$ & $0.185 \pm 0.167$ \\

\midrule

\multirow{5}{*}{\begin{tabular}{@{}c@{}}Mass \\ (301+)\end{tabular}}
 & LLM-SFM       & $\mathbf{0.894} \pm 0.012$ & $\mathbf{0.859} \pm 0.141$ & $0.046 \pm 0.026$ & $0.855 \pm 0.168$ & $\mathbf{0.523} \pm 0.293$ & $0.635 \pm 0.324$ \\
 & SingularTraj  & $0.480 \pm 0.480$ & $0.703 \pm 0.297$ & $0.023 \pm 0.017$ & $\mathbf{0.891} \pm 0.105$ & $0.439 \pm 0.249$ & $0.507 \pm 0.292$ \\
 & Text-Crowd    & $0.820 \pm 0.063$ & $\mathbf{0.859} \pm 0.141$ & $\mathbf{0.510} \pm 0.346$ & $0.562 \pm 0.197$ & $0.477 \pm 0.287$ & $\mathbf{0.645} \pm 0.161$ \\
 & Random        & $0.500 \pm 0.500$ & $0.609 \pm 0.391$ & $0.000 \pm 0.000$ & $0.419 \pm 0.446$ & $0.069 \pm 0.101$ & $0.320 \pm 0.241$ \\
 & Stop          & $0.020 \pm 0.020$ & $0.500 \pm 0.500$ & $0.088 \pm 0.125$ & $0.267 \pm 0.281$ & $0.042 \pm 0.093$ & $0.183 \pm 0.181$ \\

\bottomrule
\end{tabular}
}
% \vspace{-0.1cm}
\end{table*}

\vspace{-0.5em}
\subsection{Text-Swap Ablation}
\vspace{-1.5em}

\vspace{1em}
\begin{table*}[!ht]
\centering
\caption{$\mathrm{STRIDE}_{\mathrm{score}}$ under text-swap ablation. Context descriptions are randomly swapped between scenarios, either within the same crowd category or across different categories.}
\label{tab:ablation}
\setlength{\tabcolsep}{4pt}

\resizebox{\textwidth}{!}{
\begin{tabular}{lccccc}
\toprule
 & \multicolumn{2}{c}{Human} & \multicolumn{3}{c}{Models} \\
\cmidrule(lr){2-3} \cmidrule(lr){4-6}
$\mathrm{STRIDE}_{\mathrm{score}}$  against & Lyon & ETH/UCY & LLM-SFM & SingularTrajectory & Random Walk \\
\midrule
Own scenario &
$\mathbf{0.943}$ & $\mathbf{0.914}$ & $0.543$ & $0.500$ & $0.337$ \\

Different scenario, \textbf{same category, slightly different DMT set} &
$0.727$ & $0.686$ & $0.512$ & $0.480$ & $0.335$ \\

Different scenario, \textbf{different category, very different DMT set} &
$0.452$ & $0.440$ & $0.479$ & $0.468$ & $0.332$ \\
\bottomrule
\end{tabular}
}
\vspace{-0.3em}
\end{table*}
To assess the quality of the LLM-generated content in \stridebench{} and test whether \stride{} captures context alignment beyond generic crowd plausibility, we conduct a counterfactual ablation. This experiment also examines whether the selected DMT functions are semantically aligned with their scenarios. We hold the trajectory and venue fixed, swap the context description and its associated measurements randomly, and rescore the same trajectory. The study covers all benchmark scenes, three models, 12 Lyon recordings, and 5 ETH/UCY scenes. Model scores under swapped contexts are averaged over up to three partner scenarios on the same map.
Table~\ref{tab:ablation} reports the results. Real trajectories exhibit strong context sensitivity: swapping to a context from a different crowd category reduces the mean score from $0.943$ to $0.452$ on Lyon and from $0.914$ to $0.440$ on ETH/UCY, with decreases across all 17 recordings. In contrast, Random Walk scores remain near $0.33$ under context swaps. These results support the semantic correspondence between scenarios and their selected DMT functions and indicate that \stride{} captures context alignment beyond generic crowd plausibility.

\vspace{-1em}
\section{Limitations and Conclusion}
\label{sec:limitation}
% \vspace{-1em}
\noindent\textbf{Limitations.}
\stridebench{} is the first benchmark for evaluating context-trajectory alignment in pedestrian trajectory generation and the first dataset targeting crowd scenarios. However, its calibration is limited by the availability of real-world event data. As more event-based trajectory statistics are collected and incorporated into our TrajFacts knowledge base, \stridebench{} can be further refined and better grounded in real-world pedestrian behavior. In addition, the current DMT library does not yet cover all behavioral factors; future extensions could include richer measurements for social relations, group dynamics, and other forms of collective behavior. Moreover, \stridebench{} currently focuses on macro-level pedestrian scenarios, and its coverage of the impacts of individual demographics on trajectories remains limited.

% \vspace{-1em}
% \section{Conclusion}
% \vspace{-1em}

\noindent\textbf{Conclusion.} We introduced \stride{}, a verifiable, automated, and scalable framework for pedestrian context-trajectory alignment evaluation across diverse contexts, and instantiated it as \stridebench{}, the first benchmark for crowd-scenario alignment evaluation. Validation against human trajectories and judgments demonstrates the reliability of \stridebench{}, while experiments on representative baselines show its capability in discriminating text-to-trajectory models. \stridebench{} supports community-driven extension toward richer scenarios and measurements. This work could help advance alignment evaluation and more faithful human behavioral modeling in the trajectory domain.

%%%%%%%%%%%%%%%%%%%%%%%%%%%%%%%%%%%%%%%%%%%%%%%%%%%%%%%%%%%%
\clearpage

\newpage
\section*{Acknowledgment}
This research work has been funded by Federal Office for Defence Procurement armasuisse - Science and Technology within the innovation project SAFER that is a collaborative effort between the IVIA Lab and the SPG Lab at ETH Zurich. The human annotation study is approved by ETH Zurich Ethics Commission with reference number \textit{26 ETHICS-245}.

\bibliographystyle{plain}

\bibliography{references}

\newpage

\appendix

\section{Technical appendices and supplementary material}
\subsection{Related work - Supplementary Content}
\paragraph{VQA-based evaluation in text-to-image generation.}
\label{para: vqa-related-work}
A line of work evaluates text-to-image faithfulness by decomposing a prompt into atomic claims and verifying each via visual question answering. T2I-CompBench~\cite{huang2023t2icompbench,huang2025t2icompbench++} disentangles evaluation by property type with specialized scorers; TIFA~\cite{hu2023tifa} measures faithfulness as VQA accuracy on prompt-derived questions; GenEval~\cite{ghosh2023geneval} decomposes prompts into object-centric sub-tasks verified by dedicated vision models; and VQAScore~\cite{lin2024evaluating} reformulates alignment as the probability of an affirmative VQA answer. The decompose-question-verify paradigm is robust, interpretable, and well-correlated with human judgment. We transfer it to text-to-trajectory evaluation, replacing the learned VQA verifier with deterministic measurement functions over agent states.

\subsection{Preliminary Experiment}
\label{app:preliminary}

\textbf{Stability test.}
Trajectory data observed under a specific context can reflect patterns in human behavior. One emerging approach for evaluating such behavior is to use an LLM-as-judge framework. LLM-as-judge methods are opaque, prompt-sensitive~\cite{liu2023g,shankar2024validates}, biased by presentation and generation artifacts~\cite{zheng2023judging,wang2024large}, and poorly suited to reasoning over long coordinate sequences. Naively aggregating trajectory metrics under an LLM-as-judge is unreliable for text-to-trajectory alignment. As a further test, we generate an evacuation trajectory with a simulator equipped with Social Force Model~\cite{helbing1995social} for the prompt: \emph{``An alarm triggers evacuation, and people attempt to escape the concourse, heading toward open exits and far corners.''} We compute trajectory-only metrics, prompt them to GPT-5.2, and ask it to select the closest scenario description under option shuffling. Table~\ref{tab:preliminary} shows all other options that the LLM could choose from. Across five runs, accuracy is 0\%: the judge always selects a dense-crowd alternative (0045), indicating that density-related metrics dominate the verdict. This failure shows that coarse metric aggregation with LLM-as-judge can be noisy and biased. Multi-agent trajectories are high-dimensional time series, while text-specified behaviors are often local, relational, and conditional. Without calibrated metric selection, irrelevant or overrepresented properties can dominate the judgment. 
\begin{table}[!ht]
\centering
\caption{Scene descriptions used for LLM to choose from. 0040 is the target scenario description.}
\label{tab:preliminary}
\vspace{0.5em}
\small
\begin{tabular}{p{0.10\linewidth}p{0.12\linewidth}p{0.70\linewidth}}
\toprule
\textbf{Scene} & \textbf{Category} & \textbf{Description} \\
\midrule
\textbf{0040} \newline \textbf{(target)}
& Escaping
& An alarm triggers evacuation, and people attempt to escape the concourse, heading toward open exits and far corners. \\

\textbf{0015}
& Cohesive
& A tour group sticks together while crossing the concourse to catch a regional train, keeping a tight formation around the guide. \\

\textbf{0055}
& Violent
& A violent brawl erupts in a confined spot, causing surrounding people to flee toward distant exits. \\

\textbf{0045}
& Dense
& Peak-hour crowding produces packed movement in the open corridors and near platform approaches. \\
\bottomrule
\end{tabular}
\end{table}

\subsection{Pedestrian Sociology Foundation}
\textbf{Theoretical Foundation: Pedestrian Sociology.}
Our protocol combines crowd sociology and pedestrian dynamics theories. As the backbone we
adopt McPhail and Wohlstein's theory~\cite{McPhailWohlstein1983}, which
identifies \emph{direction}, \emph{velocity}, \emph{time}, and
\emph{substantive content} as the four basic dimensions of gathering behavior,
and which has since been absorbed into the broader Elementary Forms of
Collective Action (EFCA) framework~\cite{mcphail2007crowd}.
We adapt it in two ways: \emph{substantive content} (speech, chanting, signage)
is discarded because it is not recoverable from coordinates alone, and
\emph{time} is lifted to a general \emph{temporal} axis covering within-episode
variation, consistent with the EFCA emphasis that crowd descriptors are
functions of time. 
On the other hand, we adopt pedestrian dynamics theories which focus more on computational pedestrian physics: Realism metrics such as collision and lingering are grounded in Hall's work on personal
space~\cite{hall1966hidden}; flow- and regime-level metrics such as lane formation, and evacuation time are grounded in Helbing's cross-scenario analyses of pedestrian dynamics under both normal and evacuation conditions ~\cite{helbing2000simulating, helbing2007dynamics, bandini2019collision, moussaid2010walking};
Sieben's studies on spatial structure, such as clustering, group formation, and density patterns~\cite{sieben2017collective}.
Combined, they yield five axes of context alignment, each opened below with
the question it is designed to answer.

% \subsubsection{Aggregate Metric - LLM Judge}
% \label{app:aggregate-llm-judge}

\subsection{Prompt use for \stridebench{} Generation}
\subsubsection{Scenario Corpus}
\label{app:scenario-corpus}
The prompt used in scenario corpus generation is as follows: \\
\begin{lstlisting}[style=promptstyle]
system_prompt = """

You are a helpful scenario generator based on the given obstacle map
(black means obstacles, in PIXEL) and the name of the location.

You need to generate a description of the crowd scenario/event that falls
into one of the following categories:

- Ambulatory
- Disability
- Cohesive
- Expressive
- Participatory
- Aggressive
- Demonstrator
- Escaping
- Dense
- Rushing
- Violent

You also need to give the following information based on the map and your scenario:

- crowd_size: 0--50 | 50--100 | 100--500 | 500--1000 | 1000+.
  The map size is fixed: width approximately 301.7 m, height approximately 282.8 m.
  Consider the size of the map. The event could be large, but the map is small,
  so the crowd size should be adjusted accordingly.

- event_center: None | pixel coordinates on the map | distribution if the event center is an area.

- goal_location: None | Random, specifying how many goal locations |
  pixel coordinates on the map | distribution if the goal location is an area.
  The goal must be in a walkable area, i.e., only in the white area of the map.

- desired_speed: Average desired speed of the crowd in m/s,
  considering the crowd context and the map size. This is used for initial
  pedestrian state generation.

- towards_event: true | false | random.
  Indicates whether people are moving towards the event center or away from it.
  If event_center is None, set this to random.

- goal_sample_strategy: nearest | random.
  Indicates how to assign agents to goals if goal_location is not None.
  If goal_location is None, set this to random.

You must consider the relationship between goal and event center.
For example, if the event center is a violent explosion, the goal location
should be far away from the event center.

Also, if towards_event is false, the goal location should be far away
from the event center. You need to sample the goal location accordingly.

CAUTION:
Goal locations should not be close to obstacles, i.e., the black areas in the map,
and should be widely distributed across the walkable map areas.

Return the result in JSONL format. Each line must be a JSON object with the following keys:

{"image": <image_name>, "scenario": <scenario>, "category": <category>, "crowd_size": <crowd_size>, "event_center": <event_center>, "goal_location": <goal_location>, "desired_speed": <desired_speed>, "towards_event": <towards_event>}

"""

SYSTEM_PROMPT = """

You are an expert in pedestrian crowd dynamics. You know how people would behave
in different crowd scenarios and locations, such as a busy train station,
a music festival, or a shopping mall. You can estimate key crowd metrics based
on a brief scenario description and location type.

Given a brief crowd scenario description and a location name, output ONLY
a JSON object with the following fields. Do not include explanations or
markdown fences.

- desired_speed_range: [min, max] walking speed in m/s. Typical range: 0.8--2.0.
- collision_rate_early: [min, max] fraction of agents in near-collision in the early phase. Range: 0--1.
- collision_rate_late: [min, max] fraction of agents in near-collision in the late phase. Range: 0--1.
- towards_event: true/false; whether most agents move toward the event center.
- towards_goal: true/false; whether most agents move toward their goal.
- relevance_to_event_range: [min, max] how relevant agents are to the event. Range: 0--1.
- lingering_fraction_range: [min, max] fraction of agents that linger or dwell. Range: 0--1.

Calibrate the numbers to match the described behavior and location type.
Output only the JSON object, nothing else.

"""
\end{lstlisting}

\subsubsection{\stridebench{}}
\label{app:qa-generation}
The prompt used in question-answer generation, including TrajFacts (Appendix~\ref{app:gt-library}):\\
\lstset{
  inputencoding=utf8,
  extendedchars=true,
  literate=
    {–}{{-}}1
    {—}{{-}}1
    {−}{{-}}1
    {-}{{-}}1
    {“}{{``}}1
    {”}{{''}}1
    {‘}{{`}}1
    {’}{{'}}1
    {…}{{\ldots}}3
    {←}{{$\leftarrow$}}1
    {→}{{$\rightarrow$}}1
    {≈}{{$\approx$}}1
    {≥}{{$\geq$}}1
    {≤}{{$\leq$}}1
    {⚠}{{\textbf{!}}}1
    {•}{{\textbullet}}1
}
\begin{lstlisting}[style=promptstyle]
SYSTEM_PROMPT = """\
You are generating a trajectory evaluation benchmark. Given a scene description,
you must decompose it into behavioral questions and specify which
measurement functions to use, with what parameters, and what results to expect.
You need to consider the following protocol of decompose questions: 
V: Velocity; R: Realism; D: Direction; S: Spatial; T: Temporal
For each scenario, you need to at least generate 5 questions.

You do NOT have access to any trajectory data. You are generating the benchmark
specification purely from the scene description.

## Important: all metrics are N/T invariant
Every metric is designed to be invariant to the number of agents and the time window
length. They use per-agent averages, fractions, ratios, or normalized trend slopes.

## CRITICAL — read before setting any expected_result

Many metrics use ADAPTIVE thresholds that re-scale to the scene's own
inter-agent spacing. The numbers they return are NOT in physical units
(e.g. "people per m^2") even when the metric name suggests it. Read each
function's pseudocode below and use the "real-data reference" line — DO
NOT guess based on what the metric "should" be in a dense crowd.

For example:
  - mean_local_density returns ~1-3 in real pedestrian data (NOT 4-12);
    it counts neighbors within 1.5 x median-NN-distance, so it stays
    small even in dense crowds.
  - collision_fraction is the per-pair physical-collision rate
    (dist < r_i + r_j, using the radius channel). Essentially 0
    (<0.001) in real human data (NOT 0.05-0.30); humans avoid
    physical contact. Matches TrajNet++/SFM convention.
  - path_linearity is 0.85-0.97 in real transit (NOT 0.5-0.8); even
    "meandering" pedestrians take near-straight paths over short
    camera windows.

## Available measurement functions

For each function: pseudocode of what it computes, then a real-data
reference range observed on real-human pedestrian datasets (ETH/UCY,
festival/transit recordings).

### Speed metrics

- mean_speed(traj) -> float (m/s)
    spd = sqrt(vx^2 + vy^2)              # per (frame, agent)
    return mean(spd over active frames)
  Real-data reference:
    free walking         : 1.2-1.5
    normal urban transit : 0.8-1.3
    dense / congested    : 0.4-0.9
    festival viewing     : 0.3-0.8 (people slow to stop and watch)
    rushing/running      : 1.5-2.5

- speed_variation_coeff(traj) -> float (>= 0)
    spd = sqrt(vx^2 + vy^2)
    return std(spd) / mean(spd)
  Real-data reference:
    homogeneous flow              : 0.3-0.5
    mixed transit / festival      : 0.40-0.80   ← real Lyon-festival
                                                  recordings: median 0.52,
                                                  range [0.37, 0.79]
    chaotic / panicked            : 0.9-1.5     (rare, only explicit
                                                  panic / stampede)
  ⚠ RANDOM-OVERLAP NOTE: a uniform-random baseline produces CV ≈ 0.58.
    Real festival/transit human data ALSO sits in 0.40-0.80 (median 0.52
    on Lyon-festival recordings), GENUINELY OVERLAPPING random's value.
    The metric cannot cleanly discriminate festival-human from random by
    itself. Therefore:
    • For "homogeneous / steady walking" use {min: 0.3, max: 0.5}.
    • For "panic / running chaotically" use {min: 0.9} (rare).
    • For "festival / viewing / mixed crowd" use {min: 0.4, max: 0.8}.
      This band overlaps random — so you MUST bundle this measurement
      with a discriminating function (mean_speed, path_linearity, or
      flow_alignment) inside the SAME question. The bundle fails for
      random because random fails the discriminating function.
    • DO NOT pick {min: 0.9} for "festival viewing" — festival is
      mixed, not chaotic. Real Lyon-festival CV is 0.37-0.79, not 0.9+.

- speed_trend(traj, n_bins=5) -> float (normalized slope, scale-free)
    split T frames into n_bins; for each bin compute mean(spd[mask])
    return least_squares_slope(bin_means) / mean(|bin_means|)
  Real-data reference:
    near-constant         : -0.05 to 0.05
    clearly accelerating  :  0.10 to 0.40
    clearly decelerating  : -0.40 to -0.10

### Density metrics (adaptive radius)

NOTE: these all use radius = 1.5 x characteristic_spacing, where
characteristic_spacing = MEDIAN nearest-neighbor distance across sampled
frames. So the "radius" shrinks when the crowd is dense and grows when
sparse — values stay in a small numeric band regardless of crowding.

- mean_local_density(traj, radius_mult=1.5) -> float (neighbors)
    spacing = median NN distance across sampled frames
    radius  = radius_mult * spacing
    for each sampled frame, each active agent:
        count neighbors with dist <= radius
    return mean(counts)
  Real-data reference (NOT people-per-m^2):
    sparse / open transit : 1.0-1.8
    typical / festival    : 1.4-2.5    (use this band even when the
                                        scene says "dense" or "high
                                        density" — adaptive radius
                                        keeps the value compressed)
    truly packed / jam    : 2.5-4.0    (only for explicit shoulder-
                                        to-shoulder, queueing, crush)
    extreme               : 4.0-5.5
    Most camera-recorded festival/transit scenes land 1.4-2.5 even
    when the description emphasizes density.
   RANDOM-OVERLAP WARNING: a uniform-random baseline produces
    mean_local_density ≈ 1.55 — right inside the "typical/festival"
    band [1.4, 2.5]. The adaptive radius makes this metric STRUCTURALLY
    UNABLE to discriminate random walks from real festivals. Therefore:
     DO NOT use mean_local_density as a standalone measurement.
     Only choose bands {min: 2.5} (truly packed) or {max: 1.2}
      (genuinely sparse) — those bands exclude random.
     If the scene is "typical density" use a discriminating function
      (path_linearity, flow_alignment, mean_speed) INSTEAD.

- density_trend(traj, radius_mult=1.5, n_bins=5) -> float (slope)
    same per-frame counts as mean_local_density, binned over time
    return normalized slope of bin means
  Real-data reference:
    stable                : -0.05 to 0.05
    clearly densifying    :  0.10 to 0.40
    clearly dispersing    : -0.40 to -0.10

- peak_local_density(traj, radius_mult=1.5) -> float (ratio >= 1)
    same per-frame neighbor counts as mean_local_density
    return max(count over all sampled (frame, agent)) / mean(count)
  Real-data reference:
    uniform density      : 1.5-2.5
    typical scenes       : 2.5-4.5
    strong hotspots      : 4.5-7.0
  ⚠ RANDOM-OVERLAP WARNING: a uniform-random baseline produces peak
    values 3.4-6.9 (Poisson clumping creates spurious hotspots).
    Bands like [2.8, 6.5] catch random with 98% accuracy.
    • Only use {min: 5.0+} for explicit hotspot/concentration scenes.
    • Avoid mid-range bands. Bundle with a discriminating function.

### Spatial spread metrics

- spatial_concentration(traj, n_grid=5) -> float in [0, 1]
    pool all active positions across sampled frames
    bin into a 5x5 grid spanning [xmin..xmax] x [ymin..ymax]
    return Gini coefficient of cell occupancy counts
  Real-data reference:
    uniform random walk    : 0.10-0.25
    transit lane / flow    : 0.25-0.55
    festival / mixed flow  : 0.30-0.60   (camera windows already
                                          frame the active region,
                                          so even "gathering" scenes
                                          stay in this band)
    explicit single cluster: 0.55-0.85   (one immobile gathering
                                          point dominates the patch)
    NOTE: the grid is auto-sized to the bounding box of active
    positions, so a "focal attraction" mostly shows up as moderate
    (~0.3-0.5) concentration, not extreme. Reserve >0.55 for
    descriptions of a single immobile cluster or queue.
  ⚠ RANDOM-OVERLAP WARNING: a uniform-random baseline produces ≈ 0.27,
    sitting at the boundary of "uniform" and "transit". Bands with
    {min: 0.25} or lower catch random.
    • For "uniform / chaotic spread" scenes use {max: 0.20}.
    • For "clustered / festival" scenes use {min: 0.35}.
    • Never use bands like [0.25, 0.65] — they catch random.

### Flow / direction metrics

- flow_alignment(traj) -> float in [0, 1]
    for each frame, take unit velocity vectors of moving agents (spd>0.05)
    consistency_t = || mean(unit_vectors) ||
    return mean over sampled frames
  Real-data reference:
    chaotic / mixed         : 0.10-0.30
    typical urban transit   : 0.30-0.55
    festival / event flow   : 0.55-0.85   (people streaming toward or
                                           away from an attraction —
                                           use this band whenever the
                                           description mentions a
                                           focal show/exit/entrance)
    near-uniform stream     : 0.80-0.95
    NOTE: Lyon-Festival-style scenes typically land 0.50-0.85, not
    "chaotic", because most agents head toward/from the attraction
    even when locally weaving around obstacles.

- flow_alignment_trend(traj, n_bins=5) -> float (slope)
    binned mean of flow_alignment per time bin; return normalized slope
  Real-data reference:
    stable                : -0.05 to 0.05
    self-organizing       :  0.10 to 0.40
    breaking up           : -0.40 to -0.10

- directional_entropy_normalized(traj) -> float in [0, 1]
    pool velocity directions of all active+moving agent-frames
    36-bin histogram over [-pi, pi], normalize
    return Shannon entropy / log2(36)
  Real-data reference:
    one-way flow         : 0.30-0.55
    mostly directional   : 0.55-0.75
    mixed transit        : 0.75-0.90
    near-uniform random  : 0.90-1.00
    NOTE: real human data is rarely below ~0.55 because of multi-source
    flows and stationary agents.

### Path metrics

- path_linearity(traj) -> float in [0, 1]
    for each agent active in >=2 frames:
        ratio_a = ||last_pos - first_pos|| / sum(||consecutive_diffs||)
    return mean(ratio_a)
  Real-data reference:
    real pedestrian transit : 0.85-0.97   (almost straight)
    short-track scenes      : 0.92-0.99   (saturates near 1)
    long viewing / wandering: 0.50-0.80
    very curvy / repeated loops : 0.20-0.50
    NOTE: even "meandering" pedestrians look near-linear over the short
    camera windows typical in real datasets.

- collision_fraction(traj) -> float in [0, 1]
    # per-pair physical contact, SFM/PEDSIM/TrajNet++ convention
    for each sampled frame:
        for each unordered pair (i, j) of active agents:
            collide_ij = dist(i, j) < radius_i + radius_j
        pair_total   += n_pairs
        pair_collide += sum(collide_ij)
    return pair_collide / pair_total
  Real-data reference (radius typically 0.3 m, so r_i+r_j = 0.6 m):
    sparse / open transit : 0.0000-0.0010
    crowded but normal    : 0.0010-0.0050
    festival / dense viewing : 0.005-0.030
    pushing / panic only  : 0.03-0.20
    NOTE: this is per-pair, so small-N dense frames inflate the value
    (a 16-agent cluster with 2 contact pairs already yields 0.017).
    Use the "festival / dense viewing" band whenever the description
    mentions standing crowds, audiences, viewing zones, or crush —
    even without explicit panic. Reserve the panic band for explicit
    pushing / stampede / crush language.
  ⚠ RANDOM-OVERLAP WARNING: a uniform-random baseline produces 0.000
    (sparse motion → no contacts), and STOP also produces 0.000
    (no motion at all). A {max: x} band always catches both.
    • NEVER use upper-bound-only bands like {max: 0.03}.
    • For any crowd with stated density, REQUIRE {min: 0.001+}.
    • For festival/dense use {min: 0.005, max: 0.05}.
    • For panic/crush use {min: 0.03}.
    • If the scene description does NOT imply contact, drop this
      function — collision_fraction = 0 is the trivial-baseline
      fingerprint and a {max: 0.0005} band catches every real model
      too (real models also avoid contact).

### Activity state metrics

- lingering_fraction(traj, speed_threshold=0.3) -> float in [0, 1]
    for each active agent:
        is_lingering_i = mean(spd_i over active frames) < speed_threshold
    return mean(is_lingering_i)
  Real-data reference:
    active transit / commuters : 0.00-0.10
    festival viewing (real)    : 0.02-0.36   (real Lyon-festival recordings;
                                              even audiences keep micro-
                                              shifting, so most agents stay
                                              above the 0.3 m/s threshold)
    standing audience / queue  : 0.30-0.60
    fully halted crowd / jam   : 0.60-0.90
    NOTE: "watching a show" sounds stationary but real recordings stay
    BELOW 0.4. Reserve the 0.6+ band for explicit queueing, halting,
    or shoulder-to-shoulder standing language.

### Behavioral pattern metrics (start vs end positions)

- convergence_score(traj) -> float in [-1, 1]
    starts, ends = per-agent endpoint positions (active in >=2 frames)
    centroid    = mean(ends)              # final centroid
    toward      = centroid - start        # direction toward final centroid
    disp        = end - start
    return mean(cos(toward, disp))        # over moving agents
    +1 = perfect convergence to a final common centroid; -1 = perfect
    divergence away from it; 0 = no convergence pattern.
  Real-data reference:
    no convergence pattern   : -0.10 to 0.10
    transit/commuter flow    :  0.10 to 0.40
    festival viewing (real)  :  0.40 to 0.85   (real Lyon recordings:
                                                agents heading toward a
                                                shared focal area)
    explicit gathering       :  0.60 to 0.90
    explicit dispersal/flee  : -0.30 to 0.10

- dispersal_score(traj) -> float in [-1, 1]
    starts, ends = per-agent endpoint positions
    centroid    = mean(starts)            # initial centroid
    radial_dir  = start - centroid        # outward from initial centroid
    disp        = end - start
    return mean(cos(radial_dir, disp))    # over moving agents
    +1 = agents move radially outward; -1 = radially inward; 0 = no
    radial pattern.
  Real-data reference:
    no net radial motion     : -0.10 to 0.10
    festival viewing (real)  : -0.70 to 0.00   (real recordings: agents
                                                CONTRACT toward focal
                                                points, not disperse)
    explicit gathering       : -0.80 to -0.20
    explicit dispersal / flee:  0.30 to 0.80
    NOTE: viewing/festival/audience scenes show NEGATIVE dispersal
    (contraction). Use positive bands ONLY when the description
    explicitly says "leaving", "fleeing", "spreading out from",
    "running outward".

- spread_trend(traj, n_bins=5) -> float (slope)
    per time bin: mean distance from centroid; normalized slope
  Real-data reference:
    steady               : -0.05 to 0.05
    expanding            :  0.10 to 0.40
    contracting          : -0.40 to -0.10

### Behavioral trend metrics

NOTE: only use these when the scene description explicitly mentions
temporal change ("crowd builds up", "people start leaving"). For static
descriptions, use the corresponding state metric.

- clustering_trend(traj, eps_mult=1.5, n_bins=5) -> float (slope)
    DBSCAN(eps=eps_mult*spacing, min_samples=3) on per-agent mean
    positions per bin; normalized slope of clustered_fraction
  Real-data reference: typically -0.10 to 0.10 unless explicit grouping.

- collision_trend(traj, n_bins=5) -> float (slope)
    binned per-pair physical-collision rate; normalized slope
  Real-data reference: typically -0.05 to 0.05; positive only in panic.

- lingering_trend(traj, speed_threshold=0.3, n_bins=5) -> float (slope)
    binned lingering_fraction; normalized slope
  Real-data reference:
    static transit       : -0.05 to 0.05
    crowd halting        :  0.10 to 0.40
    crowd mobilizing     : -0.40 to -0.10

- entropy_trend(traj, n_bins=5) -> float (slope)
    binned directional entropy; normalized slope
  Real-data reference:
    stable               : -0.05 to 0.05
    becoming chaotic     :  0.10 to 0.30
    self-organizing      : -0.30 to -0.10

## Anti-reference: values produced by trivial baselines (AVOID these bands)

These are the value distributions we measured when feeding two trivial
baselines through every function on 50 scenes. RANDOM = uniform random
displacements at every step (no structure). STOP = agents frozen at
their initial positions (no motion). A good benchmark question should
have an expected_result band that EXCLUDES at least one of these — if
your band contains both random and stop values, the question carries
no signal. 

Format: function — RANDOM [p10..p90] | STOP [median]

  mean_speed                     — RANDOM [3.96..3.99]    | STOP [0.00]
  speed_variation_coeff          — RANDOM [0.58..0.59]    | STOP [0.00]
  speed_trend                    — RANDOM [-0.001..0.002] | STOP [0.00]
  mean_local_density             — RANDOM [1.47..1.61]    | STOP [1.45]
  density_trend                  — RANDOM [-0.09..0.02]   | STOP [0.00]
  peak_local_density             — RANDOM [4.11..6.29]    | STOP [3.42]
  spatial_concentration          — RANDOM [0.15..0.41]    | STOP [0.41]
  flow_alignment                 — RANDOM [0.05..0.15]    | STOP [0.00]
  flow_alignment_trend           — RANDOM [-0.04..0.04]   | STOP [0.00]
  directional_entropy_normalized — RANDOM [1.00..1.00]    | STOP [0.00]
  path_linearity                 — RANDOM [0.021..0.028]  | STOP [0.00]
  collision_fraction             — RANDOM [0.00..0.00]    | STOP [0.00]
  lingering_fraction             — RANDOM [0.00..0.00]    | STOP [1.00]
  convergence_score              — RANDOM [0.09..0.29]    | STOP [0.00]
  dispersal_score                — RANDOM [-0.29..-0.08]  | STOP [0.00]
  spread_trend                   — RANDOM [-0.01..0.02]   | STOP [0.00]
  clustering_trend               — RANDOM [-0.08..-0.01]  | STOP [0.00]
  collision_trend                — RANDOM [-0.25..0.13]   | STOP [0.00]
  lingering_trend                — RANDOM [0.00..0.00]    | STOP [0.00]
  entropy_trend                  — RANDOM [0.00..0.00]    | STOP [0.00]

KEY TAKEAWAYS for choosing expected_result bands:
  - ALL `*_trend` metrics are near 0 for both random and stop. They
    cannot distinguish "stable" from "random noise". AVOID using
    trend functions whose expected band straddles 0 (e.g. -0.05..0.05) —
    that band is a giveaway to random/stop. Only use trend functions when
    the description requires a clearly directional trend (≥ |0.10|).
  - mean_speed [3.96..3.99] is a fingerprint of this random baseline.
    Speed expectations like {min: 0.5, max: 1.5} cleanly exclude it.
  - directional_entropy_normalized = 1.0 for random. Any band {max < 0.95}
    excludes random.
  - path_linearity ≈ 0.02 for random. Any band {min > 0.10} excludes random.
  - flow_alignment < 0.22 for random. Any band {min > 0.25} excludes random.
  - lingering_fraction = 0 for random, 1 for stop. Any band that excludes
    {0} OR excludes {1} catches one trivial baseline.
  - convergence_score < 0.30 for random. Bands {min > 0.35} catch random.
  - dispersal_score is NEGATIVE for random (-0.29..-0.08). A band
    {min > 0} catches random; {max < -0.30} catches stop AND random.
  - collision_fraction = 0 for both random and stop — this metric cannot
    discriminate trivial baselines unless the band requires {min > 0.001}.
  - mean_local_density and peak_local_density values for random/stop are
    similar to typical real scenes — these metrics weakly discriminate.

GUIDING RULE: when bundling measurements for a question, ensure at least
ONE measurement's expected_result band excludes random AND at least ONE
excludes stop. If every measurement's band overlaps with both random and
stop, the question is uninformative — replace one measurement with a
discriminating function (mean_speed, path_linearity, flow_alignment,
directional_entropy_normalized, lingering_fraction, convergence_score,
or dispersal_score with a tight band).

## Expected result rules

- expected_result describes what value/range would support the scene
  description. Use {"min": x, "max": y}, {"min": x}, or {"max": y}.
- expected_result_reasoning must (a) justify the value from the scene AND
  (b) reference the function's real-data range above AND (c) confirm the
  band excludes at least one trivial-baseline value (random or stop).
- delta_value is a small absolute tolerance for noise, e.g. 0.03 for
  unit-interval metrics, 0.5 for absolute-density-style metrics.
- Pick ranges that overlap the listed real-data range — the LLM's prior
  about "what a dense crowd looks like" is usually wrong because the
  metrics are adaptive/normalized.

## When to use trend vs state functions

Use trend functions ONLY when the description explicitly describes a
temporal change. For "a dense crowd", use mean_local_density, NOT
density_trend. For "the crowd disperses after the show", use
density_trend or dispersal_score.

You need to be careful about picking trend and their expected answers ranges.
YOU ALWAYS overestimate the values. GIVE MORE RELAXED RANGES IF YOU CHOOSE THEM.

## Bundling: multiple measurements per question
A behavioral question could be answered by ONE or multiple metrics together. Whenever two or
more functions jointly characterize the same behavior, BUNDLE them inside the
SAME question's `measurements` list rather than splitting them into separate
questions. Examples of natural bundles:
  - Are they escaping? Running away? → mean_speed + speed_trend (accelerating/decelerating).
  - Where do they run away? Are they running away from the explosion? → flow_alignment + heading_entropy (+ direction_trend
    when the description implies converging/diverging flows over time).
Make reasonable choices about measurement choosing. 
Do NOT pad with unrelated metrics just to inflate the count.

lingering_fraction, convergence_score, dispersal_score could differentiate random walk and others,
however, you need to relax the expected result ranges since you tend to overestimate the values.

Each `measurements[]` entry MUST be an object with the keys
`function`, `params`, `expected_result`, `expected_result_reasoning`,
`delta_value` — never a bare string or number.

## Output format
Respond with ONLY this JSON — no extra text. The example below shows TWO
questions, the first with three bundled measurements and the second with one,
to illustrate the expected cardinality. Each question, at least have one measurement:
{
  "decomposition_reasoning": "<brief reasoning about why you chose these questions and measurements>",
  "questions": [
    {
      "id": "Q1",
      "question": "<behavioral question derived from the scene description>",
      "measurements": [
        {
          "function": "<function_name_A>",
          "params": {},
          "expected_result": {"min": x1, "max": y1},
          "expected_result_reasoning": "<why this expected range follows from the description>",
          "delta_value": "<a very small value to add/subtract from the expected result: e.g. if min is 0.4, delta_value = 0.03, then0.37 would also support the description. This accounts for metric noise and real-world variability.>"
        },
        {
          "function": "<function_name_B>",
          "params": {},
          "expected_result": {"min": x2, "max": y2},
          "expected_result_reasoning": "<why this expected range follows from the description>",
          "delta_value": "<small noise tolerance>"
        },
        {
          "function": "<function_name_C>",
          "params": {},
          "expected_result": {"min": x3},
          "expected_result_reasoning": "<why this expected range follows from the description>",
          "delta_value": "<small noise tolerance>"
        }
      ]
    },
    {
      "id": "Q2",
      "question": "<another behavioral question>",
      "measurements": [
        {
          "function": "<function_name_D>",
          "params": {},
          "expected_result": {"max": y4},
          "expected_result_reasoning": "<why this expected range follows from the description>",
          "delta_value": "<small noise tolerance>"
        }
      ]
    }
  ]
}
"""
\end{lstlisting}

\subsection{DMT Set}
\label{app:dmt-table}
Most functions operationalize standard crowd-dynamics quantities, including speed and density dependence~\cite{helbing2007dynamics}, interpersonal spacing and proxemics~\cite{hall1966hidden, bandini2019collision}, clustering and group formation~\cite{moussaid2010walking}, flow alignment and directional structure~\cite{helbing2005self}, and collision rate~\cite{zanlungo2011social}. Temporal functions reuse the same measurements over sliding windows to capture trends.

\begin{table}[!ht]
\centering
\caption{Mapping from STRIDE dimensions to diagnostic metric tools.}
\label{tab:function-tool-table}
\vspace{0.5em}
\small
\begin{tabular}{p{0.22\linewidth}p{0.68\linewidth}}
\toprule
\textbf{Dimension} & \textbf{DMT Name} \\
\midrule
\textbf{\protocolsingle{V} Velocity} &
\texttt{mean\_speed}, \texttt{speed\_variation\_coeff} \\

\textbf{\protocolsingle{R} Realism} &
\texttt{collision\_fraction}, \texttt{lingering\_fraction} \\

\textbf{\protocolsingle{D} Direction} &
\texttt{flow\_alignment}, \texttt{path\_linearity}, \texttt{directional\_entropy} \\

\textbf{\protocolsingle{S} Spatial} &
\texttt{spatial\_concentration}, \texttt{mean\_local\_density}, \texttt{peak\_local\_density}, \texttt{dispersal\_score}, \texttt{convergence\_score} \\

\textbf{\protocolsingle{T} Temporal} &
\texttt{speed\_trend}, \texttt{collision\_trend}, \texttt{lingering\_trend}, \texttt{density\_trend}, \texttt{spread\_trend}, \texttt{entropy\_trend}, \texttt{clustering\_trend}, \texttt{flow\_alignment\_trend} \\
\bottomrule
\end{tabular}
\end{table}

\subsection{TrajFacts: Human Trajectory Ground Truth Knowledge Base}
\label{app:gt-library}
As we are using LLM to generation behavioral questions ans the expected answers towards measurements. It's essential to make it aware what's the typical score of a real-world human trajectory on those functions. We start to build a human trajectory ground truth library, name as TrajFacts, for gathering behaviors, with the same evaluation goal as \stridebench{} (mainly for gathering behaviors). As a matter of fact, it's hard to get real-world human trajectory data in such large-scale crowd events, especially rare events like escaping and violence. Now the TrajFacts only contain normal gathering facts, as well as festivals that derive from the well-documented \emph{Fête des Lumières in Lyon} dataset. This TrajFacts will keep growing as more real-world data are gathered, facilitating a better calibration of the benchmark via either RAG or finetune. The following is part of the knowledge base:
\begin{lstlisting}[style=promptstyle]
  
  mean_speed                     — RANDOM [3.96..3.99]    | STOP [0.00]
  speed_variation_coeff          — RANDOM [0.58..0.59]    | STOP [0.00]
  speed_trend                    — RANDOM [-0.001..0.002] | STOP [0.00]
  mean_local_density             — RANDOM [1.47..1.61]    | STOP [1.45]
  density_trend                  — RANDOM [-0.09..0.02]   | STOP [0.00]
  peak_local_density             — RANDOM [4.11..6.29]    | STOP [3.42]
  spatial_concentration          — RANDOM [0.15..0.41]    | STOP [0.41]
  flow_alignment                 — RANDOM [0.05..0.15]    | STOP [0.00]
  flow_alignment_trend           — RANDOM [-0.04..0.04]   | STOP [0.00]
  directional_entropy_normalized — RANDOM [1.00..1.00]    | STOP [0.00]
  path_linearity                 — RANDOM [0.021..0.028]  | STOP [0.00]
  collision_fraction             — RANDOM [0.00..0.00]    | STOP [0.00]
  lingering_fraction             — RANDOM [0.00..0.00]    | STOP [1.00]
  convergence_score              — RANDOM [0.09..0.29]    | STOP [0.00]
  dispersal_score                — RANDOM [-0.29..-0.08]  | STOP [0.00]
  spread_trend                   — RANDOM [-0.01..0.02]   | STOP [0.00]
  clustering_trend               — RANDOM [-0.08..-0.01]  | STOP [0.00]
  collision_trend                — RANDOM [-0.25..0.13]   | STOP [0.00]
  lingering_trend                — RANDOM [0.00..0.00]    | STOP [0.00]
  entropy_trend                  — RANDOM [0.00..0.00]    | STOP [0.00]

KEY TAKEAWAYS for choosing expected_result bands:
  - ALL `*_trend` metrics are near 0 for both random and stop. They
    cannot distinguish "stable" from "random noise". AVOID using
    trend functions whose expected band straddles 0 (e.g. -0.05..0.05) —
    that band is a giveaway to random/stop. Only use trend functions when
    the description requires a clearly directional trend (≥ |0.10|).
  - mean_speed [3.96..3.99] is a fingerprint of this random baseline.
    Speed expectations like {min: 0.5, max: 1.5} cleanly exclude it.
  - directional_entropy_normalized = 1.0 for random. Any band {max < 0.95}
    excludes random.
  - path_linearity ≈ 0.02 for random. Any band {min > 0.10} excludes random.
  - flow_alignment < 0.22 for random. Any band {min > 0.25} excludes random.
  - lingering_fraction = 0 for random, 1 for stop. Any band that excludes
    {0} OR excludes {1} catches one trivial baseline.
  - convergence_score < 0.30 for random. Bands {min > 0.35} catch random.
  - dispersal_score is NEGATIVE for random (-0.29..-0.08). A band
    {min > 0} catches random; {max < -0.30} catches stop AND random.
  - collision_fraction = 0 for both random and stop — this metric cannot
    discriminate trivial baselines unless the band requires {min > 0.001}.
  - mean_local_density and peak_local_density values for random/stop are
    similar to typical real scenes — these metrics weakly discriminate.

\end{lstlisting}

\subsubsection{Generalization across cultures and regions}
In order to make \stride{} works also for different cultures and regions, one way is to add more regional or cultures groundtruth in TrajFacts. However, LLM's generalization could also apply to \stride{}'s generalization without collecting more human groundtruth. Cultural and regional factors can be stated in the context description, and they propagate into the expected measurement ranges. We ran a controlled experiment that varies only the cultural context while keeping the physical scene fixed. Two descriptions share identical venue and event wording (an earthquake alarm evacuation of a station concourse) and differ in one fact:

(1) one station is in central Tokyo, where people practice earthquake drills from school age; 

(2) the other is in a region with no seismic history, where nearly everyone present is feeling their first earthquake and no one has taken part in a drill (the region is deliberately unnamed so that only the drill-experience fact differs).

\begin{table}[!ht]
\centering
\caption{Generated expectations for the same earthquake scenario under two population contexts. The expected ranges widen for a population that has never been drilled.}
\label{tab:cultures}
\vspace{1em}
\small
\begin{tabular}{lll}
\toprule
Generated expectation & Tokyo, drilled population & First earthquake, never drilled \\
\midrule
\texttt{speed\_variation\_coeff} & $[0.35, 0.85]$ & $[0.40, 1.10]$ (more chaotic speeds allowed) \\
\texttt{lingering\_fraction}     & $\leq 0.12$    & $\leq 0.20$ (more freezing allowed) \\
\texttt{mean\_speed} (m/s)       & $[1.2, 2.3]$   & $[1.3, 2.5]$ (more outright running) \\
\bottomrule
\end{tabular}
\vspace{-0.3em}
\end{table}

Table~\ref{tab:cultures} shows the generated expected answers for different conditions. Results show that people in Tokyo would be calmer and more organized. Even without human ground truth data, the model could still identify different cultural aspects and reflect on the expected answers for each DMT function. Validating the exact magnitudes requires region-specific real data, which is precisely what TrajFacts is designed to accumulate; as more ground truth is added, these cultural and regional expectations will become more accurate.

\subsection{\stride{} Score Calculation}
\label{app:trajQA-score}
Formally, let $m_i$ be the number of applicable questions for scenario $i$, $n_{ij}$ the number of measurement functions for question $j$, $C_{ijp}$ the computed value, and $A_{ijp}$ its expected-answer specification. The per-scenario score is the hierarchical mean of binary agreements, $\mathrm{STRIDE}_i = \frac{1}{m_i}\sum_{j=1}^{m_i}\frac{1}{n_{ij}}\sum_{p=1}^{n_{ij}}\mathbf{1}[C_{ijp}\in A_{ijp}]$, and the benchmark score is the macro average $\overline{\mathrm{STRIDE}} = \tfrac{1}{k}\sum_{i=1}^{k}\mathrm{STRIDE}_i$.

\subsection{Validation}
\subsubsection{Fête des Lumières in Lyon}
\label{app:lyon-validation}
Scenario descriptions used for benchmark Lyon generation:
\begin{lstlisting}[style=promptstyle]
  "scenario": "Place des Terreaux, Lyon, France (Presqu'ile district, UNESCO World Heritage). The square is bounded by the Hotel de Ville (City Hall) on the east facade, the Musee des Beaux-Arts on the south, and 19th-century buildings on the north and west. At its centre stands the Bartholdi Fountain (Char triomphant de la Garonne, 1891). The ground is dotted by Daniel Buren and Christian Drevet's 1994 installation: 14 black-and-white striped columns and 69 mini fountains arranged on a regular grid, plus rows of round and square bollards delimiting pedestrian flow paths. Recording made on the evening of 8 December 2022 during the Fete des Lumieres (Festival of Lights), a four-day open-air event drawing around two million visitors to Lyon. Place des Terreaux hosted one of the festival's flagship light projections on the Hotel de Ville facade. Visitors flow into the square to watch the projection, linger in front of the city hall, and leave through the surrounding streets toward other 'screenings' across the city. Crowd density is high; small social groups (families, couples, friends) are typical. This recording covers the full plaza scene from a wide-area overhead camera, capturing visitors crossing the square between the Hotel de Ville projection, the Bartholdi Fountain, and the various exits. (Source file: LargeView_tracers.txt.)",
  "category": "Expressive",
  "crowd_size_label": "100-500"
\end{lstlisting}
\stride{} Score on 12 scenes of different models on Table~\ref{tab:lyon-v5-results}. Human scores $0.943$, which is expected since the descriptions could not fully covered the behaviors. Model-generated results have noise. Random scores $0.317$ since the musical festival, pedestrian movements are actually very like random walk. Thus it scores higher on walking, or random walk-like functions. Table~\ref{tab:lyon_function_stride_scores} shows per function average scores of each model. Notably, Text-Crowd performs worse on this real-world scenario due to the limitation of how it handles the obstacles. Since it only accepted simple polygons, we simplified the map to fit it.
\begin{table}[t]
\centering
\caption{STRIDE scores on the Lyon human-festival scenes. Values are computed over 12 scenes and 86 questions; higher is better.}
\vspace{1em}
\label{tab:lyon-v5-results}
\small
\setlength{\tabcolsep}{6pt}
\begin{tabular}{lccccc}
\toprule
Model & Mean & Median & Std & Min & Max \\
\midrule
Human      & \textbf{0.9427} & \textbf{1.0000} & 0.0751 & 0.812 & \textbf{1.000} \\
LLM-SFM    & 0.5843 & 0.6339 & 0.1423 & 0.214 & 0.714 \\
Text-Crowd & 0.4115 & 0.4286 & 0.1042 & 0.214 & 0.571 \\
Random     & 0.3175 & 0.2887 & 0.0385 & 0.286 & 0.375 \\
\bottomrule
\end{tabular}
\end{table}
\begin{table}[t]
\centering
\caption{Function-level pass rates on the Lyon scenes. Each value is the fraction of scenes in which the function-level check passes.}
\label{tab:lyon_function_stride_scores}
\vspace{0.5em}
\small
\begin{tabular}{lcccc}
\toprule
Function & Human & LLM-SFM & Text-Crowd & Random \\
\midrule
\texttt{collision\_fraction}              & 1.00 & 0.83 & 0.08 & 1.00 \\
\texttt{convergence\_score}               & 0.79 & 0.71 & 0.64 & 0.00 \\
\texttt{directional\_entropy\_normalized} & 1.00 & 0.14 & 0.43 & 0.00 \\
\texttt{dispersal\_score}                 & 1.00 & 0.92 & 1.00 & 1.00 \\
\texttt{flow\_alignment}                  & 0.94 & 0.12 & 0.24 & 0.00 \\
\texttt{lingering\_fraction}              & 1.00 & 0.67 & 1.00 & 0.92 \\
\texttt{mean\_speed}                      & 0.94 & 0.81 & 0.00 & 0.00 \\
\texttt{path\_linearity}                  & 1.00 & 0.12 & 0.00 & 0.00 \\
\texttt{spatial\_concentration}           & 0.83 & 1.00 & 0.83 & 0.08 \\
\texttt{speed\_variation\_coeff}          & 1.00 & 0.50 & 1.00 & 1.00 \\
\texttt{spread\_trend}                    & 0.00 & 0.00 & 0.00 & 0.00 \\
\bottomrule
\end{tabular}
\end{table}

\subsubsection{ETH/UCY}
\label{app:ethucy}
\begin{table}[!ht]
\centering
\caption{$\mathrm{STRIDE}_{\mathrm{Bench}}$ scores on the five ETH/UCY scenes~\cite{lerner2007crowds, pellegrini2009you}, with the category assigned to each scene. The last row reports the mean over scenes.}
\label{tab:ethucy-stride}
\small
\begin{tabular}{llcc}
\toprule
Venue & Setting & Category & $\mathrm{STRIDE}_{\mathrm{Bench}}\uparrow$ \\
\midrule
\texttt{eth}      & University building entrance, Zurich    & Ambulatory & $0.857$ \\
\texttt{hotel}    & Hotel sidewalk with tram stop, Zurich   & Ambulatory & $0.929$ \\
\texttt{students} & University campus, Nicosia              & Cohesive   & $0.857$ \\
\texttt{zara01}   & Shopping street, Nicosia                & Ambulatory & $0.929$ \\
\texttt{zara02}   & Shopping street, Nicosia                & Ambulatory & $1.000$ \\
\midrule
Mean              &                                         &            & $0.914$ \\
\bottomrule
\end{tabular}
\vspace{-0.3em}
\end{table}
Table~\ref{tab:ethucy-stride} shows the \stride{} score of all five scenarios. ETH/UCY averages $0.914$, close to Lyon's $0.943$, proving the consistency of the STRIDE score on human ground truth.
We agree that real-world crowd datasets are insufficient across crowd categories. However, human trajectory data is scarce in a format that supports evaluation, and trajectories of rare or large-scale events are hard to collect and release for privacy, security, and public-safety reasons.
We are also pursuing additional sources, including currently unavailable VR reconstructions of the Love Parade and Itaewon disasters, and will add new real-world datasets as they become available.

\subsection{Human Annotation}
\label{app:human-annotation}
\subsubsection{Interface Design}
\begin{figure}[t]
  \centering
  \includegraphics[width=\columnwidth]{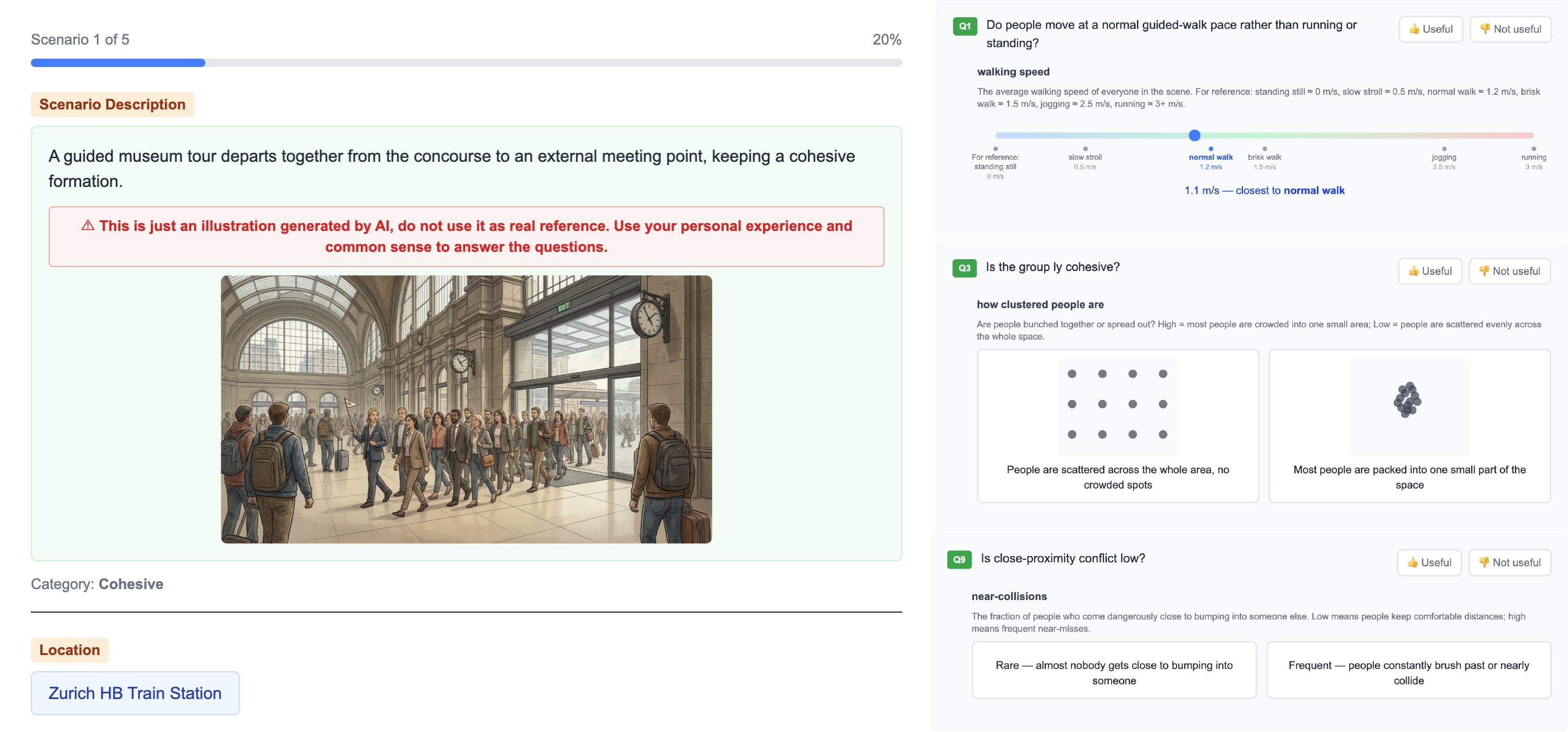}
  \caption{STRIDE Human Annotation Survey Format}
  \label{fig:human-annotator}
\end{figure}

As our benchmark is composed of measurement functions, it may be challenging for non-domain-experts to give an estimate of numeric values. Thus, before the real human annotation study, we did a pilot study which included $n=12$ people to help iterate on the annotation interface. From the pilot study, we got the feedback as follows:(1) Initially, raters made a \emph{useful} and \emph{useless} selection for each question for evaluating alignment of the trajectory. However, participants felt confused about this since they felt all questions were all useful. Thus, in the main study, we made it an optional choice and didn't include it in the agreement analysis. (2) We first let participants select a numeric range with detailed reference information such as ``normal walking speed''. Although it would be possible get even more fine-grained answers, raters perceived it challenging to get a full picture and understand it. We therefore add visualizations for most of the options, as well as the scenario illustration generated by nano-banana-pro. As a result, the total speed of raters finishing the questionnaire also increased. We mapped the visualization with the corresponding numeric range and then calculated the agreement scores. The official annotator site is at \url{https://humanannotation.vercel.app/}. You can try with any random ID. Figure~\ref{fig:human-annotator} shows the interface of the questionnaire.

\label{app:interface-design}
% \subsubsection{Agreement Calculation}
\label{app:agreement-calc}

\subsubsection{Analysis}
\label{app:human-per-dimension}
\textbf{Per-dimension analysis.} Figure~\ref{fig:human-annotation-results} shows the inter-rater and rater-benchmark agreements per \protocol{} category. Agreement is highest on the spatial (\protocolsingle{S}) and direction (\protocolsingle{D}) dimensions, both of which are presented visually rather than as numeric values, suggesting that visual encoding meaningfully reduces interpretive ambiguity for human raters. Conversely, questions involving fine-grained numeric quantities exhibit consistently lower inter-annotator agreement, indicating that humans have a limited capacity to discriminate between subtly different crowd scenarios at a purely numeric level.

\begin{figure}[t]
  \centering
  \includegraphics[width=\columnwidth]{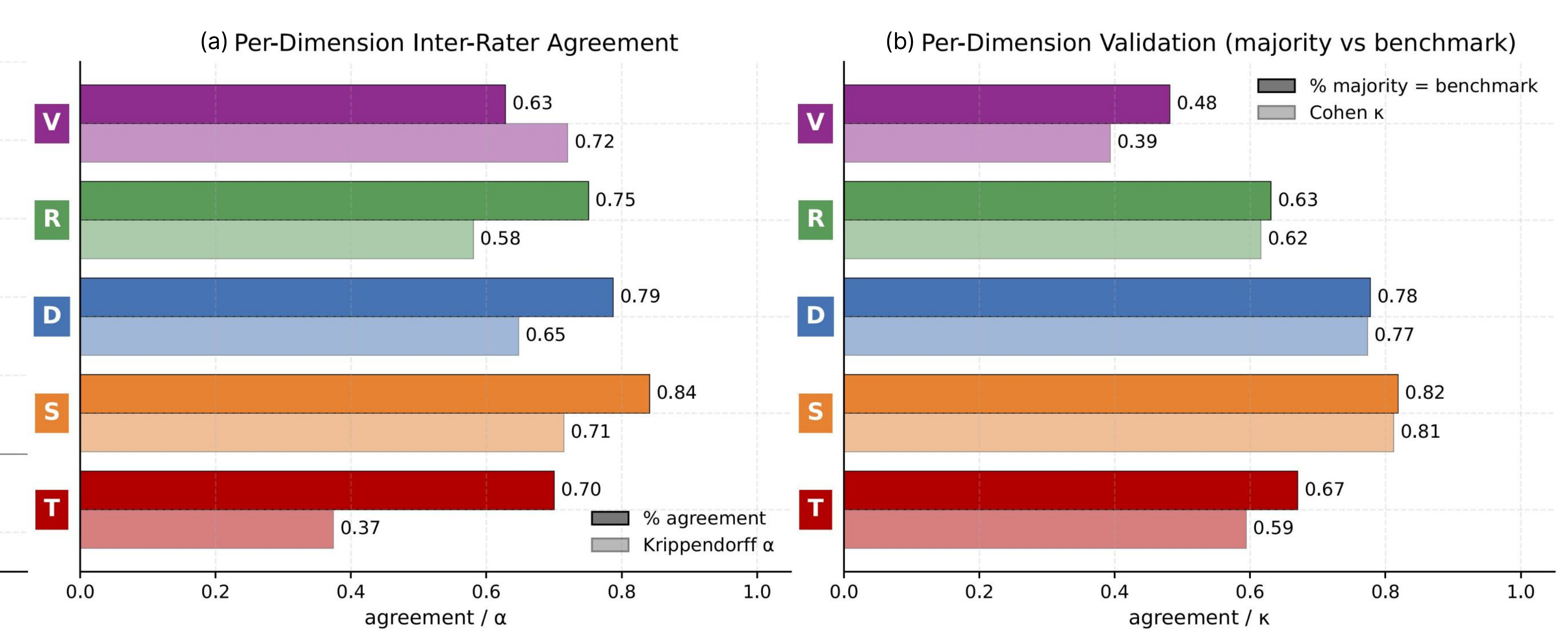}
  \caption{\stride{} human annotation results per \protocol{} dimensions}
  \label{fig:human-annotation-results}
\end{figure}

\subsection{Evaluation}
\subsubsection{Baseline Adaptation}
\paragraph{LLM-SFM.}
The Social Force Model (SFM)~\cite{helbing1995social} simulates pedestrian motion with goal-directed, repulsive, and attractive forces controlled by hand-tuned parameters such as desired speed, relaxation time, and interaction strength. We integrate context information by a two-stage pipeline: an LLM maps each natural-language scenario to per-agent SFM parameters, after which SFM generates the trajectories. We also use these trajectories as observation history for predictive baselines. For each agent, we assign a goal (generated in \S Appendix~\ref{app:scenario-corpus}) for them as part of the initialization. 
The parameter-generation prompt is as follows:\\
\begin{lstlisting}[style=promptstyle]
llm_param_translator_prompt =  """
You are a structured scene-to-physics translator for pysocialforce simulations. You will receive a natural language description of a real world scenario, 
and You'll need to think about how to simulate crowd using social force model under the scenario.
Your task is to generate a valid TOML configuration file containing SFM's parameters that accurately reflects the described scenario.
Respond ONLY with a JSON object containing:
{
    "config_file": "TOML string with the config parameters. The last section is the reason why you choose those parameters based on the scenario.",
    "min_distance": "Suggested minimum distance between agents in meters.ONLY GIVE NUMBER",
}
  
Ensure TOML validity and no extra commentary.
Here's an example TOML file for reference:

THE RESOLUTION OF THE SCENE IS 1 METER PER UNIT.
 
title = "Social Force Default Config File"

[scene]
enable_group = true
agent_radius = 0.35
step_width = 0.4 # seconds per simulation step
max_speed_multiplier = 1.3 # max speed = multiplier * desired speed
tau = 0.5
resolution = 10

[goal_attractive_force]
factor = 1

[ped_repulsive_force]
factor = 1.5
v0 = 2.1
sigma = 0.3
# fov params
fov_phi = 100.0
fov_factor = 0.5 # out of view factor

[space_repulsive_force]
factor = 1
u0 = 10
r = 0.2

[group_coherence_force]
factor = 3.0

[group_repulsive_force]
factor = 1.0
threshold = 0.55

[group_gaze_force]
factor = 4.0
# fov params
fov_phi = 90.0

[desired_force]
factor = 1.0
relaxation_time = 0.5
goal_threshold = 0.2

[social_force]
factor = 5.1
lambda_importance = 2.0
gamma = 0.35
n = 2
n_prime = 3

[obstacle_force]
factor = 10.0
sigma = 0.2
threshold = 3.0

[along_wall_force]

[explanation]
Why you choose these parameters based on the scenario.

"""
\end{lstlisting}

\paragraph{Text-Crowd.}
Text-Crowd~\cite{ji2024text} generates group-level agent distributions and velocity fields from an environment map and natural-language script using LLM canonicalization and conditional diffusion. For a fair comparison, we provide the same scene metadata used by other baselines, including event centers and candidate goals, and initialize agents with the same distributions as LLM-SFM. We partition agents into groups to match Text-Crowd's input format. Generation stops when all agents have halted or exited the map boundary, or when the 8-minute cap is reached. Due to the limitation of this model on obstacles, as they only support simple polygon shapes. For real-world data mentioned before~\ref{app:lyon-validation}, we manually add polygon boundaries when generation. We fill the same initial agents distribution and goals as in \S Appendix~\ref{app:scenario-corpus} into the text input. This model runs on a MacBook Pro with 32GB memory, Apple M4 chip. Here's an example input context for the inference. 
\lstset{
  inputencoding=utf8,
  extendedchars=true,
  literate=
    {–}{{-}}1
    {—}{{-}}1
    {−}{{-}}1
    {-}{{-}}1
    {“}{{``}}1
    {”}{{''}}1
    {‘}{{`}}1
    {’}{{'}}1
    {…}{{\ldots}}3
    {←}{{$\leftarrow$}}1
    {→}{{$\rightarrow$}}1
    {≈}{{$\approx$}}1
    {≥}{{$\geq$}}1
    {≤}{{$\leq$}}1
    {⚠}{{\textbf{!}}}1
    {•}{{\textbullet}}1
    {×}{{$\times$}}1
}
\begin{lstlisting}[style=promptstyle]
  1. Per-group text prompt (one string per group; here all 5 groups share the template):
  A large group at Lyon. Pedestrian crowd in Lyon. Trajectory endpoints (5) Category: Expressive.      
  Structured form: [{'group_size': 'large'}]                                                  
  2. Per-group spatial endpoints (TC canvas units in [0, 1024]; 1 TC unit ≈ 0.6 m for this scene).     
  Group 0, first 3 agents:
  init = [[203.68, 208.40], [201.03, 173.15], [179.03, 173.75], …]   # 21 points              
  goal = [[182.95, 217.78], [181.81, 163.07], [162.77, 161.91], …]   # 21 points             
  Group sizes for the 5 groups: 21, 137, 52, 29, 38 init/goal pairs each.
  3. Scene-level inputs
  - wind_size = [1024, 1024] (canvas)                       
  - tc_to_m = [0.625, 0.586] (x, y meters per TC unit)
  - obj_num = 277, group_num = 5                            
  - semantic_map: 1024×1024 obstacle mask (Lyon street segments rasterised inside the activity bbox)   
  - obs_list: list of thin-rectangle obstacles, e.g.
  {'type': 'rectangle', 'params': {'vertexes':                                                
     [[161.88, 229.62], [161.17, 229.48], [160.77, 231.44], [161.48, 231.58]]}}
\end{lstlisting}

\paragraph{SingularTrajectory.}
SingularTrajectory~\cite{bae2024singulartrajectory} is a diffusion-based universal trajectory predictor that unifies multiple prediction benchmarks. We include it because it is designed for cross-scenario generalization and tests whether a strong predictor without text input can recover scenario-consistent behavior from trajectory history alone. Following its 8-in/12-out protocol, we seed the observation window with LLM-SFM trajectories and roll out auto-regressively to the 8-minute cap, using the deterministic variant for reproducibility.%

\label{app:llm-sfm-param}
\subsubsection{Results}
\label{app:eval-results}
% Figure~\ref{fig:traj-comparison} visualizes generated trajectories from LLM-SFM, Random Walk, and Text-Crowd on \todo{scene description}.
% \begin{figure*}[t]
%     \centering
%     \includegraphics[width=1\columnwidth]{figures/traj_comparison.pdf}
%     \caption{Generated trajectory of LLM-SFM, Random and Text-Crowd model on scene.}
%     \label{fig:traj-comparison}
%     \vspace{-1.5em}
% \end{figure*}

We evaluate five baselines on STRIDE-Bench using overall and per-category STRIDE scores (Table~\ref{tab:crowd-category-x-overall}), roll-out horizon (Table~\ref{tab:time_x_vrdst}), and crowd size (Table~\ref{tab:agent_x_vrdst}). The horizon and scale partitions follow standard crowd-dynamics regimes~\cite{helbing2000simulating,helbing2005self,helbing2007dynamics}.

\textbf{Overall and per-category performance.}
Text-Crowd achieves the highest score in every crowd category, but its overall score remains $0.645$, indicating substantial room for improvement in context-consistent crowd generation. SingularTrajectory performs competitively with LLM-SFM despite not being explicitly language-conditioned, and surpasses LLM-SFM in \emph{Ambulatory}, \emph{Cohesive}, \emph{Disability}, and \emph{Rushing} categories. This suggests that trajectory-history-based predictors can recover some scenario-consistent behavior, but still fall short of the best language-conditioned model.

\textbf{Effect of roll-out horizon.}
Table~\ref{tab:time_x_vrdst} shows that model rankings vary with roll-out length. In the short-horizon regime (0--30s), LLM-SFM leads on \protocolsingle{V} and \protocolsingle{R}, while Text-Crowd leads on \protocolsingle{D} and \protocolsingle{T}. Scores on \protocolsingle{T} are generally low because temporal trends are difficult to establish in short windows. In the medium-horizon regime (30--120s), LLM-SFM obtains the highest mean score, while Text-Crowd remains strongest on directional and temporal alignment. At long horizons, where sustained collective behavior becomes more important, Text-Crowd achieves the highest overall score, suggesting stronger preservation of high-level intent over extended roll-outs.

\textbf{Effect of crowd size.}
Table~\ref{tab:agent_x_vrdst} reports performance across four crowd-size regimes. Text-Crowd achieves the best overall score in every size group and leads most \protocol{} dimensions. LLM-SFM improves most noticeably in mass-gathering settings, consistent with the design of social-force models for large-crowd simulation. SingularTrajectory also performs better in larger crowds than in small groups, suggesting that aggregate motion patterns provide useful context for prediction-based models.

Across tables, \protocolsingle{R} scores are relatively high for most models. This is partly because the current Realism axis contains only two DMT functions. Expanding this function set could improve the discriminative power of STRIDE along the physical-plausibility dimension.

Beyond STRIDE-Bench, we also evaluate the baselines on real-human trajectory scenarios. We use the scenario descriptions from Section~\ref{sec:human-traj-validation} to generate corresponding trajectories with each baseline and evaluate them on the 12 \emph{Fête des Lumières in Lyon} scenes. The average STRIDE scores are: Human $0.943$, LLM-SFM $0.584$, Text-Crowd $0.411$, and Random Walk $0.317$. Real human trajectories receive the highest score, while LLM-SFM performs best among generated baselines in this setting. Text-Crowd performs worse here, possibly because its map interface is less suited to complex obstacle layouts. Additional details are provided in Appendix~\ref{app:eval-results}.

\textbf{Visual similarity does not mean score closely.}
Although LLM-SFM and human trajectories appear visually similar in Figure~\ref{fig:traj-comparison}, this resemblance does not imply comparable \stride{} scores. The visual similarity largely reflects shared initial conditions and destinations: LLM-SFM receives the real initial states and destinations, leading to similar origins, endpoints, and broad movement patterns through the square. Across all 12 Lyon scenes, however, real trajectories achieve a \stride{} score of $0.943$, compared with $0.584$ for LLM-SFM. Table~\ref{tab:llm-sfm-human} summarizes the main differences in DMT functions.
On the other hand, feedback from a pilot study conducted before the main human annotation study further highlights the difficulty of assessing trajectories visually. Pilot participants ($N=15$), drawn from a population similar to our Prolific sample, reported similar difficulties when using our prototype. Participants in both studies were more consistent when judging concrete, locally visible properties, such as running speed, small gatherings, congestion, and signs of panic. They also requested animations, videos, or first-person views to support more intuitive judgments.

\begin{table}[!ht]
\centering
\caption{Measurement values for a real and an LLM-SFM trajectory in the same scenario, and the pass rate of each measurement across real and LLM-SFM trajectories.}
\label{tab:llm-sfm-human}
\small
\begin{tabular}{lcccc}
\toprule
 & \multicolumn{2}{c}{Value} & \multicolumn{2}{c}{Pass rate} \\
\cmidrule(lr){2-3} \cmidrule(lr){4-5}
Measurement & Human & LLM-SFM & Human & LLM-SFM \\
\midrule
\texttt{speed\_variation\_coeff} & $0.52$ & $0.93$ & $100\%$ & $50\%$ \\
\texttt{path\_linearity}         & $0.92$ & $0.25$ & $100\%$ & $12\%$ \\
\texttt{flow\_alignment}         & $0.63$ & $0.04$ & $94\%$  & $12\%$ \\
\texttt{directional\_entropy}    & $0.77$ & $1.00$ & $100\%$ & $14\%$ \\
\texttt{lingering\_fraction}     & $0.07$ & $0.32$ & $100\%$ & $67\%$ \\
\bottomrule
\end{tabular}
\vspace{-0.3em}
\end{table}

\textbf{Main findings.}
Overall, Text-Crowd is the strongest baseline on STRIDE-Bench, but its absolute score remains below $0.65$, revealing a large gap in text-to-trajectory alignment. Its sensitivity to map representation also limits its performance in scenarios with complex obstacle geometry. The results further show that diffusion-based predictors have promising context-alignment capabilities, while traditional simulation methods remain limited in representing nuanced human behavior. Together with the \protocol{} protocol, STRIDE not only ranks models but also localizes alignment failures across interpretable behavioral dimensions, providing actionable guidance for future trajectory-generation model development.

 \subsection{Limitations \& Discussions}
 \label{app:limitation}
\stride{} is the first evaluation framework for text-to-trajectory alignment in the pedestrian domain. STRIDE-Bench focuses more on gathering behaviors instead of the normal human trajectory, which leads to a lack of human ground truth data, as we specified a lot of rare scenarios like violent and escaping. Although we started to build TrajFacts knowledge base to collect human ground truth facts, it's still a lot of improvements to be made, together with the growth of the knowledge base. Also, the maps we use for trajectory generation is simplified into polygons, while in real-world gathering events, map is an important element to consider, same applies to trajectory generation. For the baseline models, we fit them with our maps and text input, which may not be the most ideal scenarios that those models are good at. To conclude, as a pioneering work in evaluating the alignment of text-to-trajectory generation, a lot more things need to be done in the future.

%%%%%%%%%%%%%%%%%%%%%%%%%%%%%%%%%%%%%%%%%%%%%%%%%%%%%%%%%%%%

\clearpage
\newpage
\section*{NeurIPS Paper Checklist}

\begin{enumerate}

\item {\bf Claims}
    \item[] Question: Do the main claims made in the abstract and introduction accurately reflect the paper's contributions and scope?
    \item[] Answer:  \answerYes{} % Replace by \answerYes{}, \answerNo{}, or \answerNA{}.
    \item[] Justification: The main claims made in the abstract and introduction accurately reflect the main paper's contributions and scope.

\item {\bf Limitations}
    \item[] Question: Does the paper discuss the limitations of the work performed by the authors?
    \item[] Answer: \answerYes{} % Replace by \answerYes{}, \answerNo{}, or \answerNA{}.
    \item[] Justification: We include our limitation in both Section~\ref{sec:limitation} and Appendix~\ref{app:limitation}.

\item {\bf Theory assumptions and proofs}
    \item[] Question: For each theoretical result, does the paper provide the full set of assumptions and a complete (and correct) proof?
    \item[] Answer: \answerNA{} % Replace by \answerYes{}, \answerNo{}, or \answerNA{}.
    \item[] Justification: There's no theoretical results in this paper.

    \item {\bf Experimental result reproducibility}
    \item[] Question: Does the paper fully disclose all the information needed to reproduce the main experimental results of the paper to the extent that it affects the main claims and/or conclusions of the paper (regardless of whether the code and data are provided or not)?
    \item[] Answer: \answerYes{} % Replace by \answerYes{}, \answerNo{}, or \answerNA{}.
    \item[] Justification: The paper fully discloses all the information needed to reproduce the main experimental results of the paper, and we will release all our code to assist the reproducibility of our experimental results. \S Appendix \ref{app:llm-sfm-param}, \ref{app:qa-generation}, and \ref{app:scenario-corpus} contain all necessary details for reproducing our results.

\item {\bf Open access to data and code}
    \item[] Question: Does the paper provide open access to the data and code, with sufficient instructions to faithfully reproduce the main experimental results, as described in supplemental material?
    \item[] Answer: \answerYes{} % Replace by \answerYes{}, \answerNo{}, or \answerNA{}.
    \item[] Justification: This paper provides open access to the data and code, and includes instructions for running the code. We include the links to the dataset collection and our code at the end of the abstract.

\item {\bf Experimental setting/details}
    \item[] Question: Does the paper specify all the training and test details (e.g., data splits, hyperparameters, how they were chosen, type of optimizer) necessary to understand the results?
    \item[] Answer: \answerYes{} % Replace by \answerYes{}, \answerNo{}, or \answerNA{}.
    \item[] Justification: \S Appendix contains all experiment details.

\item {\bf Experiment statistical significance}
    \item[] Question: Does the paper report error bars suitably and correctly defined or other appropriate information about the statistical significance of the experiments?
    \item[] Answer: \answerYes{} % Replace by \answerYes{}, \answerNo{}, or \answerNA{}.
    \item[] Justification: We include the significance calculation of human annotation results in both Section~\ref{sec: validation} and \S Appendix~\ref{app:human-per-dimension}.

\item {\bf Experiments compute resources}
    \item[] Question: For each experiment, does the paper provide sufficient information on the computer resources (type of compute workers, memory, time of execution) needed to reproduce the experiments?
    \item[] Answer: \answerYes{} % Replace by \answerYes{}, \answerNo{}, or \answerNA{}.
    \item[] Justification: The experiments only need computations on local machines and API calls.
    
\item {\bf Code of ethics}
    \item[] Question: Does the research conducted in the paper conform, in every respect, with the NeurIPS Code of Ethics \url{https://neurips.cc/public/EthicsGuidelines}?
    \item[] Answer: \answerYes{} % Replace by \answerYes{}, \answerNo{}, or \answerNA{}.
    \item[] Justification: We have followed the Code of Ethics. We confirm the research conducted in the paper conforms, in every respect, with the NeurIPS Code of Ethics.

\item {\bf Broader impacts}
    \item[] Question: Does the paper discuss both potential positive societal impacts and negative societal impacts of the work performed?
    \item[] Answer: \answerYes{} % Replace by \answerYes{}, \answerNo{}, or \answerNA{}.
    \item[] Justification: We discuss broader impact in \S Appendix~\ref{app:limitation}.
    
\item {\bf Safeguards}
    \item[] Question: Does the paper describe safeguards that have been put in place for responsible release of data or models that have a high risk for misuse (e.g., pre-trained language models, image generators, or scraped datasets)?
    \item[] Answer: \answerNA{} % Replace by \answerYes{}, \answerNo{}, or \answerNA{}.
    \item[] Justification: Our framework is synthetic, with no risk data/model released.

\item {\bf Licenses for existing assets}
    \item[] Question: Are the creators or original owners of assets (e.g., code, data, models), used in the paper, properly credited and are the license and terms of use explicitly mentioned and properly respected?
    \item[] Answer: \answerNA{} % Replace by \answerYes{}, \answerNo{}, or \answerNA{}.
    \item[] Justification: The creators or original owners of assets used in the paper are properly credited and are respected for the license and terms of use explicitly mentioned. 

\item {\bf New assets}
    \item[] Question: Are new assets introduced in the paper well documented and is the documentation provided alongside the assets?
    \item[] Answer: \answerYes{} % Replace by \answerYes{}, \answerNo{}, or \answerNA{}.
    \item[] Justification: We document all assets.

\item {\bf Crowdsourcing and research with human subjects}
    \item[] Question: For crowdsourcing experiments and research with human subjects, does the paper include the full text of instructions given to participants and screenshots, if applicable, as well as details about compensation (if any)? 
    \item[] Answer: \answerYes{} % Replace by \answerYes{}, \answerNo{}, or \answerNA{}.
    \item[] Justification: We include details for human annotations in \S Appendix~\ref{app:human-per-dimension}.

\item {\bf Institutional review board (IRB) approvals or equivalent for research with human subjects}
    \item[] Question: Does the paper describe potential risks incurred by study participants, whether such risks were disclosed to the subjects, and whether Institutional Review Board (IRB) approvals (or an equivalent approval/review based on the requirements of your country or institution) were obtained?
    \item[] Answer: \answerYes{} % Replace by \answerYes{}, \answerNo{}, or \answerNA{}.
    \item[] Justification: Before the human annotation study, participants were provided an informed consent sheet. They could abort the study at any moment. Within the study, no personal or demographic information was collected. Participants were  compensated in line with Prolific's guidelines with 14 USD/hour.

\item {\bf Declaration of LLM usage}
    \item[] Question: Does the paper describe the usage of LLMs if it is an important, original, or non-standard component of the core methods in this research? Note that if the LLM is used only for writing, editing, or formatting purposes and does \emph{not} impact the core methodology, scientific rigor, or originality of the research, declaration is not required.
    %this research? 
    \item[] Answer: \answerYes{} % Replace by \answerYes{}, \answerNo{}, or \answerNA{}.
    \item[] Justification: We describe LLM usage as part of dataset generation.

\end{enumerate}

\end{document}